\documentclass[sigconf]{acmart}

\usepackage{booktabs}
\usepackage{graphicx}
\usepackage{makecell}
\usepackage{multirow}
\usepackage{pifont}
\usepackage{tabularx}
\usepackage{xcolor}
\usepackage{enumitem}
\usepackage{rotating}

\setcopyright{none}
\renewcommand{\footnotetextcopyrightpermission}[1]{}

\newcommand{\cmark}{\ding{51}}
\newcommand{\xmark}{\ding{55}}

\AtBeginDocument{%
  }

\begin{document}
\settopmatter{authorsperrow=3}
\title{ReMoE: An Interpretable Mixture-of-Experts Framework for Human Mobility Recovery Forecasting Across Disasters}

\author{Yilin Lyu}
\affiliation{%
  \department{School of AI and Liberal Arts}
  \institution{Beijing Normal-Hong Kong Baptist University}
  \city{Zhuhai}
  \country{China}
}
\affiliation{%
  \institution{Shenzhen University of Advanced Technology}
  \city{Shenzhen}
  \country{China}
}
\email{lyuyilin@bnbu.edu.cn}

\author{Kang Liu}
\correspondingauthor
\affiliation{%
  \institution{Shenzhen Institutes of Advanced Technology,
               Chinese Academy of Sciences}
  \city{Shenzhen}
  \country{China}
}
\affiliation{%
  \institution{Shenzhen University of Advanced Technology}
  \city{Shenzhen}
  \country{China}
}
\email{kang.liu@siat.ac.cn}

\author{Wei Wang}
\affiliation{%
  \department{Department of Mechanical Engineering}
  \institution{City University of Hong Kong}
  \city{Hong Kong}
  \country{China}
}
\email{wwang326@cityu.edu.hk}

\author{Qifan Zhang}
\affiliation{%
  \institution{Shenzhen Institutes of Advanced Technology,
               Chinese Academy of Sciences}
  \city{Shenzhen}
  \country{China}
}
\email{622025330022@smail.nju.edu.cn}

\author{Ling Yin}
\affiliation{%
  \institution{Shenzhen Institutes of Advanced Technology,
               Chinese Academy of Sciences}
  \city{Shenzhen}
  \country{China}
}
\email{yinling@siat.ac.cn}

\author{Yushi Bai}
\affiliation{%
  \institution{Shenzhen Institutes of Advanced Technology,
               Chinese Academy of Sciences}
  \city{Shenzhen}
  \country{China}
}
\email{3250009371@student.must.edu.mo}

\renewcommand{\shortauthors}{Lyu and Liu}

\begin{abstract}
Human mobility often declines sharply following disasters and subsequently exhibits complex recovery dynamics. Effective forecasting of this recovery is critical for timely assessment and resource allocation. However, existing approaches are often developed for individual events and operate as black boxes, limiting their transferability and interpretability beyond the events they were trained on. Recent studies have identified recurring post-disaster mobility recovery types across disasters, suggesting that recovery knowledge may be transferable. Motivated by this, we propose ReMoE, an interpretable mixture-of-experts framework for cross-disaster mobility recovery forecasting. ReMoE uses a feature-guided router to adaptively combine specialized experts that learn these heterogeneous yet reusable recovery types, producing a region-specific recovery pattern. It further refines this pattern through residual correction informed by forecasts of future hazard conditions. We evaluate ReMoE on a large-scale mobility dataset covering 18 disaster events spanning hurricanes, floods, winter storms, and wildfires. ReMoE is competitive in-distribution and achieves the strongest performance on unseen events and hazard types, with larger gains under greater distribution shifts. Further analyses show that ReMoE yields interpretable learning outcomes and maintains robust performance under uncertainty. Broadly, ReMoE offers a transferable and traceable forecaster that can support anticipatory post-disaster assessment. The source code is available at: https://github.com/lvyl9909/ReMoEKDD27.

\end{abstract}



\maketitle

\section{Introduction}
\label{sec:introduction}

\begin{figure}[!ht]
\includegraphics[width=\columnwidth]{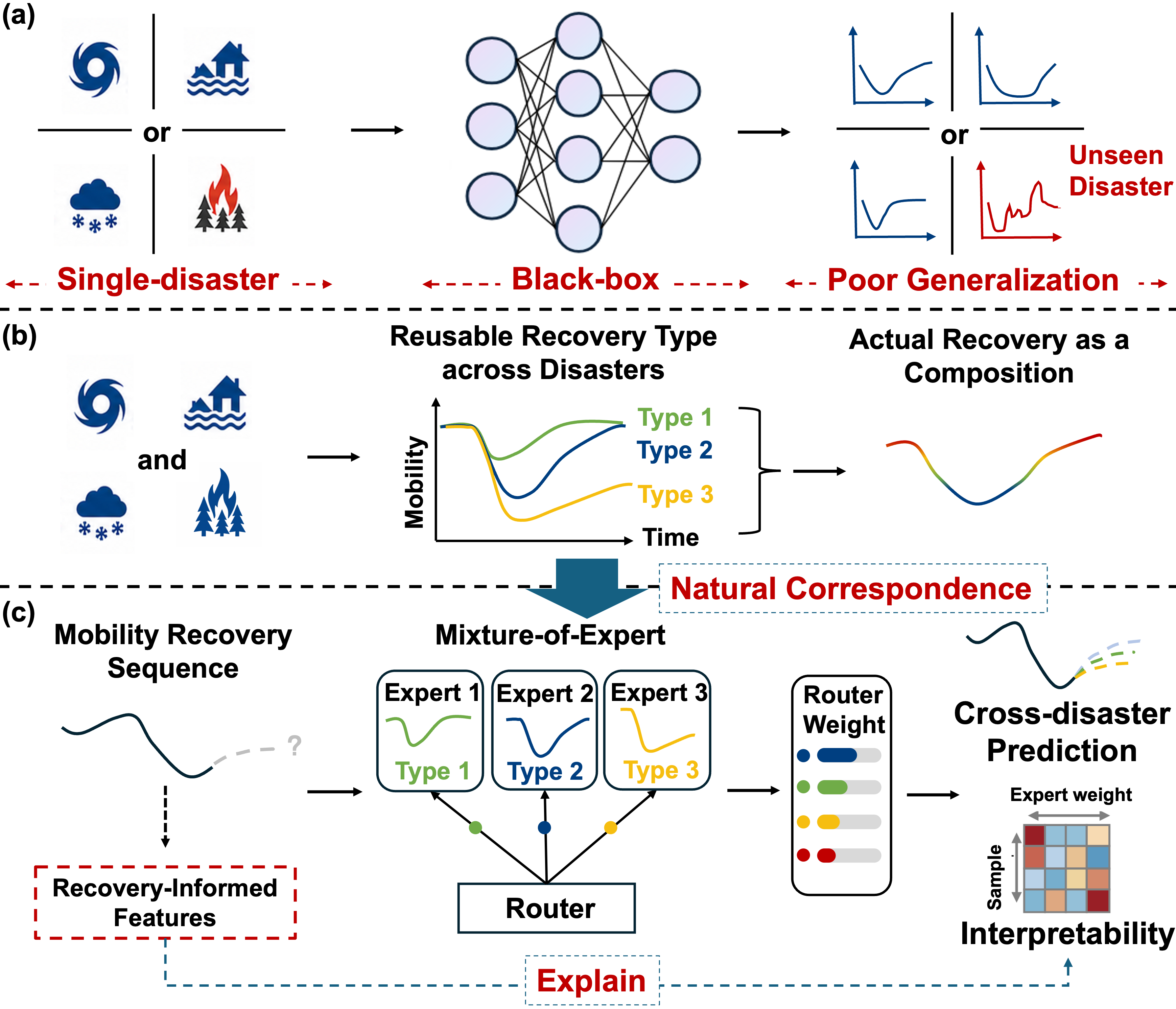}
\caption{Motivation of ReMoE.
(a) Event-specific black-box models generalize poorly to unseen disasters.
(b) Prior studies reveal recurring and compositional recovery dynamics.
(c) ReMoE learns and combines reusable dynamics for interpretable cross-disaster forecasting.}
\label{fig:intro-motivation}
\end{figure}

Human mobility undergoes a decline after disasters and subsequently exhibits a complex recovery trajectory. This process provides a timely view of how urban functions are affected and restored, while revealing substantial spatial heterogeneity in recovery across communities \cite{intro02}. However, retrospective observation is only part of the problem. In rapidly evolving scenarios, proactive decision-making requires an estimate of how mobility will evolve over the next several days. Thus, accurately predicting this dynamic recovery process holds paramount importance for informing post-disaster recovery strategies, including prioritizing resources for communities expected to experience prolonged disruption \cite{intro01}.

However, post-disaster mobility recovery forecasting faces two long-standing challenges. \textbf{First, cross-disaster generalization remains limited}. Recovery dynamics vary widely in their nature, magnitude, and duration from event to event, in large part because the events themselves differ so greatly. In addition, the heterogeneity of mobility change can vary at different geographies, manifested by the economy, demography, and physical and social infrastructure \cite{intro02-01}. Consequently, many existing studies develop ad hoc models for a single disaster type or study area \cite{m10, m8, m9}. Although informative in their settings, these models often rely on event- or city-specific correlations that may not generalize to unseen disasters or cities, limiting rapid deployment without enough local calibration data. \textbf{Second, model interpretability remains limited}. Although a growing body of research has achieved high performance, less attention has been paid to explaining model predictions and grounding them in underlying recovery mechanisms \cite{m12, m13}. However, such transparency is critical for decision support, as decision-makers must understand and evaluate a forecast rather than rely on a black-box output \cite{intro02-02}.

Recent post hoc studies have shown that mobility recovery can follow diverse but recurring patterns across disasters \cite{lr-05, intro03-01, intro-03-cluster}. For example, \citet{intro03-01} identified three archetypes from more than 2,000 empirical resilience curves, suggesting that different disasters may share three reusable recovery types. Beyond such discrete types, the actual regional recovery trajectory aggregates heterogeneous local processes and may therefore be better represented as flexible compositions of multiple latent recovery types \cite{lr-05, intro-recovery-composition}. At the same time, mixture-of-experts (MoE) models have demonstrated strong performance in learning heterogeneous data representations. They employ specialized ``experts'' to capture heterogeneous dynamics and adaptively combine their outputs through a routing mechanism \cite{intro03-03}. This compositional paradigm may be naturally aligned with regional recovery heterogeneity; e.g., specialized experts can capture reusable recovery types, while router weights combine them to represent region-specific recovery processes. This correspondence may consequently enable recovery knowledge to transfer to unseen disasters.

Motivated by these findings, we propose to leverage MoE architecture as a natural fit for predicting heterogeneous recovery dynamics. Yet, the black-box nature of standard MoE models obscures how input shapes expert specialization, making semantically grounded generalization difficult \cite{intro-interpretability01, intro-interpretability02}. To address this limitation, we propose \textbf{ReMoE}, an interpretable mixture-of-experts framework for forecasting human mobility recovery across disasters. ReMoE consists of two complementary modules. First, the Interpretable Recovery-Type Composer (IRTC) provides the main source of interpretability and cross-disaster transfer. It uses engineered features derived from early mobility observations and regional context to guide routing, making expert allocation traceable and selective. Also, its trend-residual expert design separates underlying recovery patterns from local fluctuations to promote stable expert specialization. Second, the Spatiotemporal Pressure-Informed Residual Correction (STP-RC) module incorporates forecast hazard conditions as exogenous signals to improve model performance. The main contributions can be summarized as follows:


\begin{itemize}

    \item To the best of our knowledge, we are the first to forecast transferable human mobility recovery across unseen events and hazard types. The idea of leveraging MoE to learn and compose heterogeneous yet reusable recovery dynamics supports this generalization.

    \item We develop an interpretable MoE framework that integrates a feature-guided sparse router with trend-residual experts, enabling post-hoc explanations of routing decisions and intrinsic interpretation of expert specialization.

    \item We curate a large-scale mobility dataset covering 18 disasters, over 219 million visit records, and 5,031 affected spatial units. Extensive experiments demonstrate ReMoE's superior accuracy, interpretability, and robustness, particularly under unseen-event and unseen-hazard-type settings.

\end{itemize}




\section{Preliminaries}
\label{sec:preliminaries}

\noindent\textbf{Definition 2.1 (Normalized Region-Wise Mobility Sequence).}
For disaster event $e$, let
$\mathcal{R}_e=\{r_1,\ldots,r_N\}$ denote its $N$ affected regions,
$\mathcal{T}_e=\{\tau_1,\ldots,\tau_T\}$ denote $T$ consecutive daily
intervals, and $\tau_0^e$ denote the event date. Let $f_{ij,e}(\tau)$ be
the number of directed movements from region $r_i$ to region $r_j$ on day
$\tau$. The total inter-region mobility of $r_i$ and its normalized value
are
\begin{equation}
m_{i,e}(\tau)
=
\sum_{j\neq i}
\left[
f_{ji,e}(\tau)+f_{ij,e}(\tau)
\right],
\qquad
\widetilde m_{i,e}(\tau)
=
\frac{m_{i,e}(\tau)}
{\overline m^{\mathrm{pre}}_{i,e}}.
\end{equation}
Here, $\overline m^{\mathrm{pre}}_{i,e}$ denotes the pre-event baseline, such that $\widetilde m_{i,e}(\tau)=1$ represents normal mobility. The corresponding normalized sequence over \(\mathcal{T}_e\) is
$\widetilde{\mathbf m}_{i,e}
=[\widetilde m_{i,e}(\tau_1),\ldots,
\widetilde m_{i,e}(\tau_T)]\in\mathbb{R}^{T}$.

In this study, each region corresponds to a U.S. Census tract; intra-region visits \(f_{ii,e}(\tau)\) are excluded, and origin--destination interactions are not explicitly modeled. Moreover, the normalized mobility is processed using a three-day causal moving average:
\begin{equation}
\widetilde m^{\mathrm{sm}}_{i,e}(\tau)
=
\frac{
\widetilde m_{i,e}(\tau)
+
\widetilde m_{i,e}(\tau-1)
+
\widetilde m_{i,e}(\tau-2)
}{3}.
\end{equation}
Our model input is derived from this smoothed sequence by restricting it to
a specified observation window. Let $T_{\mathrm{pre}}$ and $T_{\mathrm{post}}$ denote the numbers of
observed days before and after $\tau_0^e$, respectively. The resulting input sequence is
\begin{equation}
\widetilde{\mathbf m}^{\mathrm{in}}_{i,e}
=
\left[
\widetilde m^{\mathrm{sm}}_{i,e}
(\tau_0^e-T_{\mathrm{pre}}),
\ldots,
\widetilde m^{\mathrm{sm}}_{i,e}
(\tau_0^e+T_{\mathrm{post}})
\right].
\end{equation}

\noindent\textbf{Definition 2.2 (Spatiotemporal Pressure).}
Following the Spatiotemporal (ST) Decay Model proposed by Li et al.~\cite{Li2022}, we characterize the external hazard pressure in
the region \(r_i\) as
\begin{equation}
P_{i,e}(\tau)
=
k_e(\tau)
\sum_{j\in\mathcal{N}(i)\cup\{i\}}
w_{ij}H_{j,e}(\tau),
\end{equation}
where \(H_{j,e}(\tau)\) denotes the hazard intensity of region \(r_j\),
\(w_{ij}\) captures neighborhood influence, and
\(k_e(\tau)=k_0e^{-\alpha(\tau-\tau_0^e)}\) captures temporal decay after the event. We map this quantity to a bounded spatiotemporal pressure (STP) score:
\begin{equation}
\label{eq:stp}
\operatorname{STP}_{i,e}(\tau)
=
\frac{P_{i,e}(\tau)}
{1+P_{i,e}(\tau)}
\in[0,1).
\end{equation}
A larger STP indicates a stronger external hazard pressure. In this study, observed STP is calculated from hazard conditions within the observation window, whereas the future STP sequence \(\mathbf{STP}^{\mathrm{future}}_{i,e}\)
is derived from hazard forecasts available at the forecast origin.


\noindent\textbf{Problem Formulation.}
For each affected region \(r_i\) under disaster event \(e\), given its
observed normalized mobility sequence
\(\widetilde{\mathbf m}^{\mathrm{in}}_{i,e}\), contextual features
\(\mathbf C_{i,e}\), and forecast spatiotemporal pressure
\(\mathbf{STP}^{\mathrm{future}}_{i,e}\), we aim to predict its subsequent
\(H\)-day mobility recovery:
\begin{equation}
\widehat{\mathbf m}^{\mathrm{future}}_{i,e}
=
\mathcal F_{\theta}
\left(
\widetilde{\mathbf m}^{\mathrm{in}}_{i,e},
\mathbf C_{i,e},
\mathbf{STP}^{\mathrm{future}}_{i,e}
\right).
\end{equation}
The forecasting function \(\mathcal F_{\theta}\) is learned jointly across multiple events and hazard types to support generalization to unseen disasters.

\begin{figure*}[t]\centering\includegraphics[width=\linewidth]{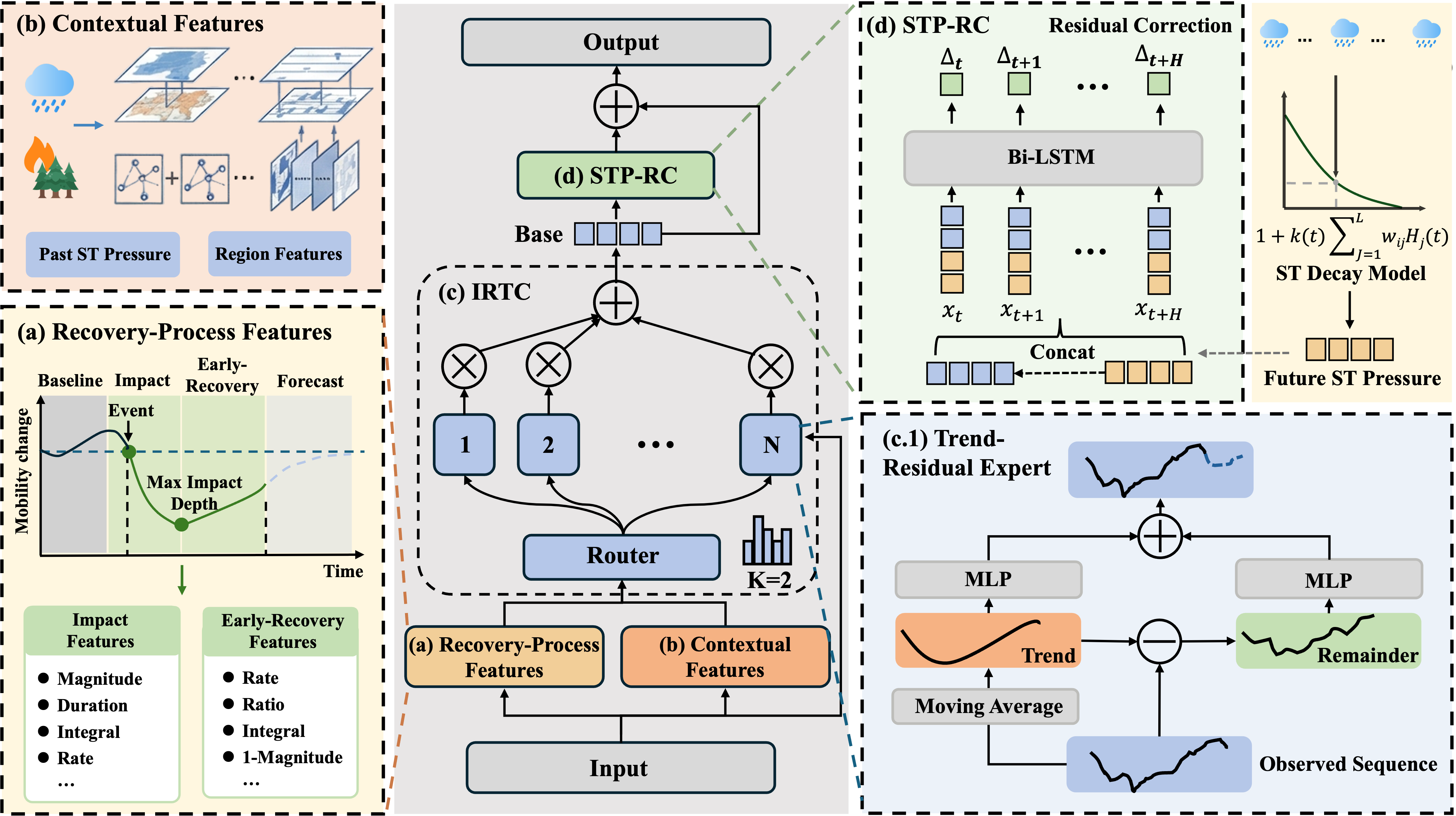}\caption{\textbf{Overview of the ReMoE framework.}}\Description{Diagram of the ReMoE framework. Mobility-derived recovery features and hazard-pressure and regional features enter a top-2 router, which combines forecasts from trend-residual experts into a Base trajectory. A BiLSTM then uses the Base trajectory and Future ST Pressure to produce a residual correction and the final mobility recovery forecast.}\label{fig:framework}\end{figure*}

\section{Methodology}
\label{sec:methodology}
\subsection{Overview}
\label{sec:methodology-overview}



The architecture of the ReMoE framework is shown in
Figure~\ref{fig:framework}. It comprises two complementary modules that integrate process-informed knowledge with data-driven learning: (1) the Interpretable Recovery-Type Composer (IRTC) and (2) the Spatiotemporal Pressure-Informed Residual Correction (STP-RC). IRTC implements an MoE architecture with a feature-guided sparse router
and a semantics-oriented expert design to learn a base recovery pattern while encouraging experts to develop distinct and stable semantic roles, thereby enhancing model interpretability. STP-RC then uses the forecast hazard conditions to estimate residual corrections to this base trajectory and produce the final forecast. Together, the two modules provide distinct yet complementary capabilities.

\subsection{Interpretable Recovery-Type Composer}
\label{sec:IRTC}

\subsubsection{\textbf{Feature-Guided Sparse Router.}}
A growing body of work suggests that MoE performance depends not only on the capacity of experts, but also on how effectively the router induces meaningful specialization among them \cite{method-3.2.1-02}. Unlike conventional MoE routers operating on latent representations, our router uses explicit features which have been shown to be informative for distinguishing subsequent recovery types \cite{intro02, method-3.2.1-informative01}, encouraging clearer expert differentiation and making routing decisions traceable \cite{method3.2.1-01}. We further adopt sparse Top-\(K\) routing strategy, which assigns each sample to only a small subset of experts and reduces excessive functional overlap among them.

From the observed normalized mobility sequence and contextual conditions, we construct 16 features in two categories: recovery-process features and contextual features. Following prior resilience-curve studies \cite{method-3.2-01, method-3.2-02}, these recovery-process features characterize properties such as magnitude, duration, integral, rate, and stability. We derive them separately from the impact and early-recovery periods because these stages provide complementary signals for subsequent recovery \cite{intro02}. Contextual features distinguish similar mobility patterns using observed STP and regional conditions (e.g., social vulnerability, population density, built-up share, and network centrality). After redundancy analysis, the final set contains five impact, six recovery, and five contextual features:
\begin{equation}
\mathbf F_{i,e}
=
\left[
\Phi^{\mathrm{imp}}
\!\left(\widetilde{\mathbf m}^{\mathrm{in}}_{i,e}\right),
\Phi^{\mathrm{rec}}
\!\left(\widetilde{\mathbf m}^{\mathrm{in}}_{i,e}\right),
\mathbf C_{i,e}
\right]
\in\mathbb R^{16},
\end{equation}
where $\Phi^{\mathrm{imp}}(\cdot)\in\mathbb R^{5}$,
$\Phi^{\mathrm{rec}}(\cdot)\in\mathbb R^{6}$, and
$\mathbf C_{i,e}\in\mathbb R^{5}$ denote the impact, recovery, and contextual feature categories, respectively. Definitions and filtering details are provided in the Appendix~\ref{app:router-features}.

The expert weights \(g_{i,e,j}\) are obtained through a softmax gating
mechanism followed by top-\(K\) selection and renormalization:
\begin{equation}
g_{i,e,j}
=
\begin{cases}
\displaystyle
\frac{\widetilde g_{i,e,j}}
{\sum_{\ell\in\mathcal K_{i,e}}\widetilde g_{i,e,\ell}},
& j\in\mathcal K_{i,e},\\[6pt]
0, & \text{otherwise},
\end{cases}
\end{equation}
where \(\widetilde g_{i,e,j}\) is the dense gating probability for
expert \(j\), and \(\mathcal K_{i,e}\) contains the indices of its
top-\(K\) probabilities. Here, we set \(K=2\).


\subsubsection{\textbf{Trend-Residual Experts.}}
Post-disaster mobility recovery is heterogeneous, with affected sub-areas exhibiting distinct recovery processes \cite{intro03-01, lr-05}. A multi-expert structure is therefore employed to capture these diverse dynamics. Ideally, each expert should specialize in a single recovery regime, allowing the weighted expert composition to provide semantic interpretability \cite{method-3.2.2-yang}. However, when experts operate directly on raw mobility sequences, long-term recovery
shapes and short-term fluctuations are entangled. Consequently, different experts may learn different mixtures of these components, obscuring their functional roles. 

To promote clearer specialization, we explicitly decompose the input to each expert into a moving-average trend and a remainder:

\begin{equation}
\mathbf{Trend}_{i,e}
=
\operatorname{MA}_{L}
\left(\widetilde{\mathbf m}^{\mathrm{in}}_{i,e}\right),
\qquad
\mathbf{Residual}_{i,e}
=
\widetilde{\mathbf m}^{\mathrm{in}}_{i,e}
-
\mathbf{Trend}_{i,e},
\end{equation}
where \(L=3\) is the moving-average window. Expert \(k\) then produces an \(H\)-day trajectory:
\begin{equation}
\mathbf{Expert}^{(k)}_{i,e}
=
\mathcal E_k^{\mathrm{trend}}
\left(\mathbf{Trend}_{i,e}\right)
+
\sigma(a_k)
\mathcal E_k^{\mathrm{res}}
\left(\mathbf{Residual}_{i,e}\right)
\in\mathbb R^{H},
\end{equation}
where the two components are MLPs, \(\sigma(a_k)\) controls the
residual contribution. The final IRTC output for region \(i\) is a weighted sum:
\begin{equation}
\mathbf{Base}_{i,e}
=
\sum_{k=1}^{E}
g_{i,e,k}\mathbf{Expert}^{(k)}_{i,e}.
\end{equation}

Note that \(\mathbf {Base}_{i,e}\in\mathbb R^{H}\) is a learned representation of basic recovery pattern  whose length matches the prediction horizon. It is subsequently forwarded to STP-RC for correction.

\subsection{Spatiotemporal Pressure-Informed Residual Correction}
\label{sec:stp-rc}

IRTC captures a basic recovery pattern from the observed sequences, but foreseeable future hazard conditions may cause additional deviations \cite{method3.3-01}. To include these influences, we employ a recurrent residual corrector to translate the evolving forecast STP into adjustments to the \textbf{Base}. Let \(\mathbf{STP}^{\mathrm{future}}_{i,e}
=[STP^{\mathrm{future}}_{i,e}(1),\ldots,
STP^{\mathrm{future}}_{i,e}(H)]^{\top}\) be the future STP sequence. It is worth noting that \(\mathbf{STP}^{\mathrm{future}}_{i,e}\) is constructed from hazard forecasts issued no later than the forecast origin and converted into tract-level STP values. It is concatenated with the $\mathbf{Base}_{i,e}$ as
\begin{equation}
\mathbf X_{i,e}
=
\operatorname{Concat}
\left[
\operatorname{sg}\!\left(\mathbf{Base}_{i,e}\right),
\mathbf {STP}^{\mathrm{future}}_{i,e}
\right]
\in\mathbb R^{H\times 2},
\end{equation}
where \(\operatorname{sg}(\cdot)\) denotes the stop-gradient operation, allowing STP-RC to use the \textbf{Base} without propagating correction gradients back to the experts, therefore IRTC remains trained because its output directly contributes to the final prediction \cite{method3.3-02}. A bidirectional LSTM then processes \(\mathbf X_{i,e}\) and estimates the pressure-driven residual:
\begin{equation}
\boldsymbol{\Delta}_{i,e}
=
\sigma(a_f)
\left[
\operatorname{BiLSTM}(\mathbf X_{i,e})\mathbf w_f+b_f
\right],
\end{equation}
where $\mathbf w_f\in\mathbb R^{2d_h}$ and $b_f\in\mathbb R$
define a shared linear projection that maps each bidirectional hidden
state to a scalar correction. \(\sigma(a_f)\) controls the overall correction magnitude. The final prediction is
\begin{equation}
\widehat{\mathbf m}^{\mathrm{future}}_{i,e}
=
\mathbf{Base}_{i,e}
+
\boldsymbol{\Delta}_{i,e}.
\end{equation}

STP-RC sequentially adjusts the \textbf{Base} to evolving forecast hazard conditions, improving predictive accuracy.


\subsection{Model Training}
\label{sec:model-training}

ReMoE is trained end-to-end by minimizing a loss function that balances prediction accuracy, router utilization and expert diversity. First, let
\(\mathcal B\subset\mathcal D_{\mathrm{train}}\) denote a mini-batch of
tract--event samples. For each sample \((i,e)\in\mathcal B\), let
\(\widetilde{\mathbf m}^{\mathrm{future}}_{i,e}\in\mathbb R^{H}\)
denote the ground-truth normalized mobility trajectory and
\(\widehat{\mathbf m}^{\mathrm{future}}_{i,e}\in\mathbb R^{H}\)
denote the corresponding ReMoE output. The trajectory fidelity loss is
\begin{equation}
\mathcal L_{\mathrm{pred}}
=
\frac{1}{|\mathcal B|H}
\sum_{(i,e)\in\mathcal B}
\left\|
\widehat{\mathbf m}^{\mathrm{future}}_{i,e}
-
\widetilde{\mathbf m}^{\mathrm{future}}_{i,e}
\right\|_2^2.
\end{equation}
Because
\(\widehat{\mathbf m}^{\mathrm{future}}_{i,e}
=
\mathbf{Base}_{i,e}
+
\boldsymbol{\Delta}_{i,e}\),
this objective jointly trains IRTC and STP-RC without requiring a
separate correction loss.

Second, MoE may suffer from ``expert collapse'', where one expert is repeatedly selected. We therefore calculate the mean router weight assigned to expert \(k\) within each mini-batch:
\begin{equation}
A_k
=
\frac{1}{|\mathcal B|}
\sum_{(i,e)\in\mathcal B}
g_{i,e,k},
\qquad k=1,\ldots,E.
\end{equation}
Here, \(g_{i,e,k}\) and \(E\) is the expert weight and the number of experts, respectively. Balanced usage is encouraged by penalizing the coefficient of variation across experts:
\begin{equation}
\mathcal L_{\mathrm{router}}
=
\frac{
\operatorname{Var}(A_1,\ldots,A_E)
}{
\operatorname{Mean}(A_1,\ldots,A_E)^2+\epsilon
},
\end{equation}
where \(\epsilon\) is a positive constant for numerical stability.

Third, to discourage experts from learning similar recovery dynamics, we calculate their batch-mean candidate trajectories:
\begin{equation}
\overline{\mathbf z}_k
=
\frac{1}{|\mathcal B|}
\sum_{(i,e)\in\mathcal B}
\mathbf{Expert}^{(k)}_{i,e}.
\end{equation}

We then compare the
mean-centered shapes of their batch-averaged trajectories and penalize
pairwise cosine similarities above margin:
\begin{equation}
\mathcal L_{\mathrm{div}}
=
\frac{2}{E(E-1)}
\sum_{1\leq p<q\leq E}
\left[
\max\left(
0,
\operatorname{sim}_{\cos}
\left(
\operatorname{ctr}(\overline{\mathbf z}_p),
\operatorname{ctr}(\overline{\mathbf z}_q)
\right)
-\rho
\right)
\right]^2,
\end{equation}
where \(\operatorname{ctr}(\cdot)\) removes the temporal mean,
\(\operatorname{sim}_{\cos}(\cdot,\cdot)\) denotes cosine similarity,
and \(\rho\) is the maximum similarity allowed without penalty.

The total training objective is
\begin{equation}
\mathcal L_{\mathrm{total}}
=
\mathcal L_{\mathrm{pred}}
+
\lambda_{\mathrm r}\mathcal L_{\mathrm{router}}
+
\lambda_{\mathrm d}\mathcal L_{\mathrm{div}},
\end{equation}
where \(\lambda_{\mathrm r}\) and \(\lambda_{\mathrm d}\) weight router balancing and expert diversity.


\section{Experiments}
\label{sec:experiments}

We evaluate ReMoE through the following evaluation questions:

\begin{itemize}[leftmargin=*, itemsep=2pt, topsep=2pt]
    \item \textbf{EQ1:} How accurately does ReMoE forecast post-disaster
    mobility recovery, and how well does it generalize to unseen events
    and hazard types? \(\rightarrow\) Sec.~\ref{sec:model-performance}

    \item \textbf{EQ2:} How does ReMoE provide interpretable insights into its learned experts and routing behavior? \(\rightarrow\) Sec.~\ref{sec:interpretability}

    \item \textbf{EQ3:} How robust is ReMoE under uncertainty and different model settings? \(\rightarrow\) Sec.~\ref{sec:uncertainty}
\end{itemize}

\subsection{Experimental Setup}
\label{sec:experimental-setup}

\subsubsection{\textbf{Datasets.}}
To support cross-disaster forecasting and evaluate generalization to unseen hazards, we construct a multi-disaster mobility dataset from over 219 million visit records provided by Veraset, a company that aggregates anonymized location data from multiple apps on Apple and Android platforms. The dataset contains 18 events between 2019 and 2025, including eight hurricanes, two floods, four winter storms, and four wildfires, covering 5031 affected census tracts. Appendix~\ref{app:dataset_description} shows the details of datasets.


Using this dataset, we evaluate ReMoE under three settings with progressively increasing distribution shift. The \textbf{in-distribution (ID)} setting randomly partitions tract--event samples while retaining all events and hazard types across the data splits. The other two settings evaluate \textbf{out-of-distribution (OOD)} generalization. First, to evaluate generalization to unseen events, we construct eight stratified \textbf{leave-one-event-out (LOEO)} folds, with one event from each hazard type withheld for testing in each fold and the remaining 14 events used for training. For hazard types with fewer than eight events, test events are reused across folds following a balanced rotation. Second, to test transferability to an unseen hazard type, we further introduce a four-fold \textbf{leave-one-hazard-type-out (LOHTO)} setting, where all events of one hazard type are withheld in each fold. Results are aggregated across folds.

\subsubsection{\textbf{Experiment Setting.}}
We use the event-aligned mobility sequence from \(t_0-9\) to \(t_0+5\) as the observation window to forecast the subsequent seven-day recovery trajectory from \(t_0+6\) to \(t_0+12\) (see Appendix \ref{app:temporal window design} for details). For ID evaluation, census tracts are partitioned into training, validation, and test sets at a ratio of 7:2:1, whereas LOEO
and LOHTO use the fixed event-level folds described above. To mitigate event-size imbalance, each training sample is weighted inversely by the number of tracts in its event. ReMoE uses a two-layer MLP router and five experts. The model is trained end-to-end with AdamW using a learning rate of \(1.2\times10^{-3}\), weight decay of \(2\times10^{-4}\), batch size 128, and random seed 42. Models are selected according to validation RMSE, and all baselines use the same observation and forecasting horizons.

\subsubsection{\textbf{Evaluation Metrics.}}
\label{subsubsec:evalaution metrics}
We evaluate forecasts with three standard metrics, RMSE, MAE, and \(R^2\). However, pointwise errors can be similar for forecasts that recover early or late, so we introduce two recovery-semantics metrics: the cumulative deficit error (CDE) measures the cumulative error over the prediction horizon, whereas the recovery timing error (RTE) measures the error in when the mobility deficit is concentrated, reported in days. For a mobility sequence \(\mathbf m=(m_1,\ldots,m_H)\), we define the total deficit and its temporal centroid as
\begin{equation}
D(\mathbf m)=\sum_{h=1}^{H}\max(1-m_h,0),
\qquad
C(\mathbf m)=\frac{\sum_{h=1}^{H}h\max(1-m_h,0)}{D(\mathbf m)+\epsilon},
\end{equation}
where \(\epsilon\) is a small positive constant for numerical stability.

To prevent large events from dominating the evaluation, all metrics are event-balanced. Let \(\mathcal E\) denote the set of evaluated events and \(N_e\) the number of test tracts in event \(e\). Each tract--event sample receives the weight
\begin{equation}
w_{i,e}=\frac{1}{|\mathcal E|N_e},
\end{equation}
so that all events contribute equally. RMSE, MAE, and \(R^2\) are computed with these weights, and
\begin{equation}
\operatorname{CDE}=\sum_{e\in\mathcal E}\sum_{i=1}^{N_e}w_{i,e}\left|D\!\left(\widehat{\mathbf m}^{\mathrm{future}}_{i,e}\right)-D\!\left(\widetilde{\mathbf m}^{\mathrm{future}}_{i,e}\right)\right|,
\end{equation}
\begin{equation}
\operatorname{RTE}=\sum_{e\in\mathcal E}\sum_{i=1}^{N_e}w_{i,e}\left|C\!\left(\widehat{\mathbf m}^{\mathrm{future}}_{i,e}\right)-C\!\left(\widetilde{\mathbf m}^{\mathrm{future}}_{i,e}\right)\right|.
\end{equation}
Lower values are better for all metrics except \(R^2\).

\subsubsection{\textbf{Baselines and variants.}}
We compare ReMoE against twelve baselines spanning four families: statistical methods (ARIMAX~\cite{experiment4.1-baseline-arima}), machine learning (XGBoost~\cite{experiment4.1-baseline-xgboost}, LightGBM~\cite{experiment4.1-baseline-lightgbm}), deep learning (FC-LSTM~\cite{experiment4.1-baseline-lstm}, TCN~\cite{experiment4.1-baseline-tcn}, DLinear~\cite{experiment4.1-baseline-dlinear}, Informer~\cite{experiment4.1-baseline-informer}, PatchTST~\cite{experiment4.1-baseline-patchtst}, iTransformer~\cite{experiment4.1-baseline-itransformer}, DUET~\cite{experiment4.1-baseline-duet}, Vanilla MoE), and a pretrained time-series foundation model (Chronos-2~\cite{experiment4.1-baseline-chronos}). For fair comparison, all models use the same time windows and the same inputs, including the observed 15-day mobility sequence, the contextual variables, and the future STP. ReMoE's 16 router features are computed from these same inputs. Our code repository details the training settings and hyperparameter configurations of all models.


\subsection{Main Performance (EQ1)}
\label{sec:model-performance}

\begin{table*}[t]
\centering
\small
\setlength{\tabcolsep}{1.8pt}
\renewcommand{\arraystretch}{1.18}
\caption{Model performance under in-distribution (ID), leave-one-event-out (LOEO), and leave-one-hazard-type-out (LOHTO) settings. Best results are in bold and second-best results are underlined. Event-level results are reported in
Appendix~\ref{app:event-level-ood-performance}.}
\label{tab:main_performance}
\newcolumntype{R}{>{\centering\arraybackslash}m{4.55em}}
\newcommand{\remoescore}[2]{\makecell[c]{#1\\[-1pt]\textcolor{gray}{#2}}}
\resizebox{\textwidth}{!}{%
\begin{tabular}{c|ccc|RR|ccc|RR|ccc|RR}
\toprule
\multicolumn{1}{c|}{\textbf{Setting}} & \multicolumn{5}{c|}{\textbf{In-distribution}} & \multicolumn{5}{c|}{\textbf{LOEO}} & \multicolumn{5}{c}{\textbf{LOHTO}} \\
\cmidrule(lr){1-1}\cmidrule(lr){2-6}\cmidrule(lr){7-11}\cmidrule(lr){12-16}
\multicolumn{1}{c|}{\textbf{Metric group}} & \multicolumn{3}{c|}{\textbf{Prediction accuracy}} & \multicolumn{2}{c|}{\textbf{Recovery semantics}} & \multicolumn{3}{c|}{\textbf{Prediction accuracy}} & \multicolumn{2}{c|}{\textbf{Recovery semantics}} & \multicolumn{3}{c|}{\textbf{Prediction accuracy}} & \multicolumn{2}{c}{\textbf{Recovery semantics}} \\
\cmidrule(lr){1-1}\cmidrule(lr){2-4}\cmidrule(lr){5-6}\cmidrule(lr){7-9}\cmidrule(lr){10-11}\cmidrule(lr){12-14}\cmidrule(lr){15-16}
\textbf{Method} & RMSE$\downarrow$ & MAE$\downarrow$ & $R^2\uparrow$ & CDE$\downarrow$ & RTE$\downarrow$ & RMSE$\downarrow$ & MAE$\downarrow$ & $R^2\uparrow$ & CDE$\downarrow$ & RTE$\downarrow$ & RMSE$\downarrow$ & MAE$\downarrow$ & $R^2\uparrow$ & CDE$\downarrow$ & RTE$\downarrow$ \\
\midrule
ARIMAX & 0.099 & 0.078 & 0.423 & 0.356 & 1.355 & 0.112 & 0.086 & 0.307 & 0.478 & 1.342 & 0.128 & 0.098 & 0.229 & 0.579 & 1.557 \\
XGBoost & 0.067 & 0.051 & 0.737 & 0.210 & 1.039 & 0.089 & 0.071 & 0.548 & 0.378 & 1.175 & 0.103 & 0.082 & 0.479 & 0.461 & 1.370 \\
LightGBM & 0.064 & 0.048 & 0.761 & 0.197 & 0.989 & 0.087 & 0.069 & 0.567 & 0.366 & 1.145 & 0.101 & 0.080 & 0.495 & 0.448 & 1.344 \\
DLinear & 0.073 & 0.051 & 0.704 & 0.216 & 1.052 & 0.088 & 0.069 & 0.555 & 0.375 & 1.127 & 0.102 & 0.079 & 0.501 & 0.442 & 1.315 \\
FC-LSTM & 0.066 & 0.049 & 0.742 & 0.205 & 1.035 & 0.084 & 0.065 & 0.594 & 0.351 & 1.095 & 0.099 & 0.078 & 0.525 & 0.426 & 1.289 \\
Informer & 0.066 & 0.048 & 0.744 & 0.197 & 1.025 & 0.091 & 0.072 & 0.505 & 0.395 & 1.197 & 0.109 & 0.086 & 0.413 & 0.486 & 1.414 \\
iTransformer & \underline{0.057} & \textbf{0.041} & \underline{0.812} & \underline{0.165} & \textbf{0.963} & 0.085 & 0.067 & 0.588 & 0.357 & 1.114 & 0.102 & 0.081 & 0.485 & 0.444 & 1.325 \\
PatchTST & 0.062 & 0.045 & 0.776 & 0.187 & 1.019 & 0.084 & 0.064 & 0.628 & 0.326 & 1.086 & 0.095 & 0.078 & 0.545 & 0.412 & 1.302 \\
TCN & 0.058 & \underline{0.042} & 0.803 & 0.170 & 0.982 & \underline{0.079} & 0.062 & 0.635 & 0.333 & \underline{1.065} & 0.097 & 0.076 & 0.565 & 0.421 & 1.279 \\
DUET & 0.059 & 0.044 & 0.797 & 0.212 & 1.048 & \underline{0.079} & \underline{0.059} & \underline{0.644} & \underline{0.304} & 1.110 & \underline{0.091} & \underline{0.071} & \underline{0.568} & \underline{0.405} & \underline{1.275} \\
Vanilla MoE & 0.068 & 0.052 & 0.740 & 0.234 & 1.011 & 0.087 & 0.066 & 0.605 & 0.326 & 1.083 & 0.095 & 0.075 & 0.530 & 0.438 & 1.290 \\
Chronos-2 & 0.069 & 0.049 & 0.733 & 0.215 & 1.067 & 0.084 & 0.068 & 0.589 & 0.368 & 1.130 & 0.102 & 0.080 & 0.486 & 0.455 & 1.352 \\
\midrule
ReMoE & \remoescore{\textbf{0.055}}{(+3.5\%)} & \remoescore{\underline{0.042}}{(-2.4\%)} & \remoescore{\textbf{0.826}}{(+0.014)} & \remoescore{\textbf{0.156}}{(+5.5\%)} & \remoescore{\underline{0.975}}{(-1.2\%)} & \remoescore{\textbf{0.073}}{(+7.6\%)} & \remoescore{\textbf{0.058}}{(+1.7\%)} & \remoescore{\textbf{0.675}}{(+0.031)} & \remoescore{\textbf{0.286}}{(+5.9\%)} & \remoescore{\textbf{0.983}}{(+7.7\%)} & \remoescore{\textbf{0.083}}{(+8.8\%)} & \remoescore{\textbf{0.069}}{(+2.8\%)} & \remoescore{\textbf{0.625}}{(+0.057)} & \remoescore{\textbf{0.335}}{(+17.3\%)} & \remoescore{\textbf{1.082}}{(+15.1\%)} \\
\midrule
w/o router features
& 0.058 & 0.043 & 0.807 & 0.171 & 1.007
& 0.077 & 0.061 & 0.645 & 0.311 & 1.035
& 0.093 & 0.074 & 0.569 & 0.393 & 1.236 \\
w/o MoE
& 0.061 & 0.046 & 0.791 & 0.182 & 1.041
& 0.083 & 0.065 & 0.595 & 0.344 & 1.125
& 0.101 & 0.079 & 0.515 & 0.425 & 1.269 \\
w/o STP-RC
& 0.064 & 0.049 & 0.785 & 0.191 & 1.025
& 0.079 & 0.062 & 0.631 & 0.352 & 1.075
& 0.094 & 0.075 & 0.555 & 0.447 & 1.214 \\
\bottomrule
\end{tabular}%
}
\end{table*}
\subsubsection{\textbf{In-distribution performance.}}
We first evaluate ReMoE in the in-distribution setting. These results establish a baseline for assessing its subsequent OOD generalization capability. As shown in Table~\ref{tab:main_performance}, ReMoE achieved the best results on three of the five metrics, with an RMSE of \(0.055\), an \(R^2\) of \(0.826\), and a CDE of \(0.156\), reducing RMSE and CDE by \(3.5\%\) and \(5.5\%\), respectively, while improving \(R^2\) by \(0.014\) compared with the strongest baseline. Its MAE of \(0.042\) matched that of TCN and was only \(0.001\) higher than iTransformer, while its RTE of \(0.975\) was within \(0.012\) days of the best-performing model. Notably, although ReMoE is designed to improve cross-disaster generalization, it also shows competitive results under in-distribution conditions.

\subsubsection{\textbf{Cross-disaster generalization.}}
ReMoE's advantage became more pronounced as the distribution shift increased. Under LOEO, which evaluates transfer to unseen events of the same hazard types, ReMoE ranked first across all five metrics, achieving an RMSE of \(0.073\), an MAE of \(0.058\), and an \(R^2\) of \(0.675\). Compared with the best baseline, it reduces errors by up to \(7.7\%\) and improves \(R^2\) by \(0.031\). Under the more challenging LOHTO setting, ReMoE maintained the best performance across all metrics, reducing RMSE and MAE by \(8.8\%\) and \(2.8\%\) and improving \(R^2\) by \(0.057\) over the strongest baselines. Interestingly, the gains are particularly evident in the recovery-semantic metrics, where ReMoE reduces CDE and RTE by \(17.3\%\) and \(15.1\%\), respectively. Among the baselines, DUET generally performs best under both OOD settings, while TCN is also competitive under LOEO. The widening gap from LOEO to LOHTO suggests that the recovery patterns ReMoE learns transfer better to unseen hazard types than those of generic forecasters.

\subsubsection{\textbf{Ablation study.}}
Ablation results are reported in Table~\ref{tab:main_performance}. We evaluate three variants: \textit{w/o router features}, which replaces the 11 engineered mobility features with a learned encoding of the observed mobility sequence, while retaining the contextual features; \textit{w/o MoE}, which replaces the five trend-residual experts with a single MLP; and \textit{w/o STP-RC}, which directly uses the \textbf{Base} as the final forecast. Our results show that: (1) removing the router features causes the mildest degradation across all settings; (2) replacing MoE with a single MLP causes substantially greater degradation, with the gap widening under stronger distribution shifts, highlighting the importance of multi-expert composition for generalization; and (3) removing STP-RC has the largest impact under ID, increasing RMSE from \(0.055\) to \(0.064\) and reducing \(R^2\) from \(0.826\) to \(0.785\), but its contribution grows less consistently under OOD settings. This may be because the relationship between hazard pressure and mobility correction varies across unseen disasters, making it harder for STP-RC
to learn a stable mapping.


\subsection{Model Interpretability (EQ2)}
\label{sec:interpretability}

\begin{figure*}[t]
\centering
\includegraphics[width=\linewidth]{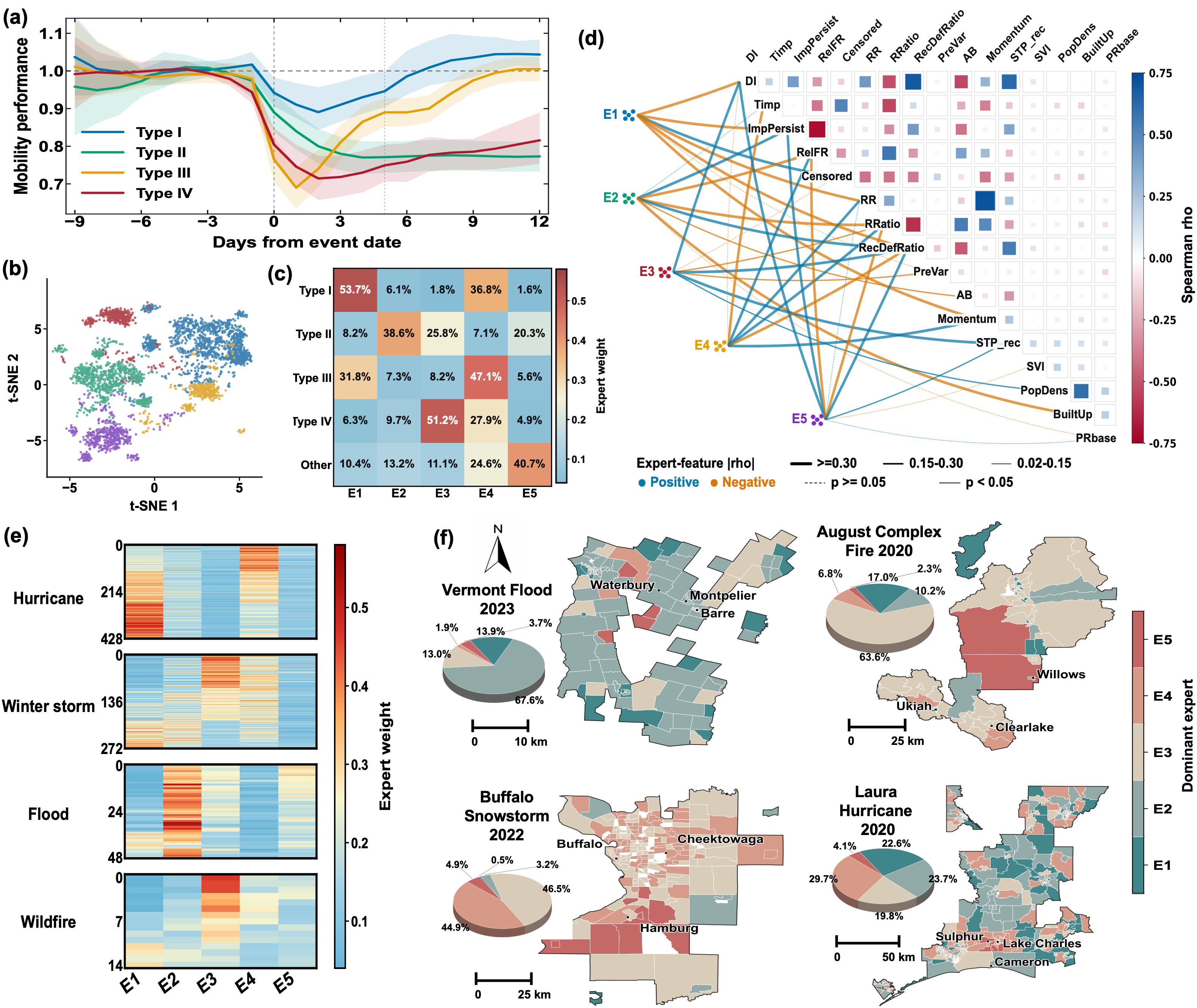}
\caption{\textbf{Model interpretability.}
(a) Four empirical recovery types identified through post-hoc clustering.
(b) t-SNE projection of router weights colored by above recovery types and ``Other'' (purple).
(c) Mean expert weights by recovery type, revealing soft specialization.
(d) Spearman associations between router features and expert weights.
(e) Expert-weight compositions across hazard types.
(f) Spatial distributions of dominant experts in representative events.}
\Description{Six-panel interpretability analysis showing empirical recovery types, router-weight representations, recovery-type specialization, feature--routing associations, cross-hazard expert reuse, and spatial heterogeneity in dominant experts.}
\label{fig:interpretability}
\end{figure*}

\subsubsection{\textbf{Understanding Learned Expert Specialization.}}
Here, we examine how expert specialization is associated with distinct recovery types. For post hoc validation, we cluster complete recovery curves from 18 disaster events and identified four representative types as shown in Figure~\ref{fig:interpretability}(a). Type I shows mild impact and rapid recovery; Type II captures prolonged disruption; Type III exhibits bimodal recovery after a pronounced impact; and Type IV represents severe impact with slow, incomplete recovery. Each sample was assigned to the nearest cluster centroid, while poorly matched samples were categorized as ``Other.'' Detailed clustering procedures and recovery-type descriptions are provided in the Appendix \ref{app:post-hoc type}.

In Figure~\ref{fig:interpretability}(b), we project the final learned routing-weight vectors using t-SNE and color each sample by its post-hoc recovery type. The result exhibits recognizable clusters, indicating that similar recovery types are associated with similar expert-weight combinations. A t-SNE comparison of the routing spaces learned by a standard MoE and ReMoE is provided in the Appendix~\ref{app:routing-space-comparison}. Building on this observation, we then quantify the correspondence between recovery types and experts by calculating the mean expert weight within each type, as shown in Figure~\ref{fig:interpretability}(c). It is observed that Type I samples are predominantly
associated with E1 (\(53.7\%\)), while Types II, III, and IV receive their highest weights for E2 (\(38.6\%\)), E4 (\(47.1\%\)), and E3 (\(51.2\%\)), respectively. E5 maintains the highest average weight for ``other'' (\(40.7\%\)), potentially reflecting its statistical association with irregular recovery behaviors. These associations provide evidence that our framework effectively learns specialized experts aligned with distinct recovery dynamics.

\subsubsection{\textbf{Interpreting Expert Routing through Features.}}
Another source of ReMoE's interpretability comes from post hoc analysis of the explicit routing features, which reveals how input characteristics influence expert selection. Figure~\ref{fig:interpretability}(d) presents the Spearman correlations between routing features and expert weights. The lower part shows that the experts exhibit distinct feature-association profiles. E1 favors less-impacted samples, as indicated by its negative association with impact depth. E2 and E3 respond more to persistent or severe disruption, with E3 also associated with higher observed STP. E4 is primarily related to recovery rate and momentum, whereas E5 is associated with greater persistence and recovery deficit but a lower restored ratio. The upper part shows correlations among routing features, providing context for their associations with expert weights.

We further conducted a global SHAP analysis to capture nonlinear feature contributions to router activations. As reported in the Appendix~\ref{app:shap-router-attribution}, impact depth, momentum, and builtup share dominate most to router activation. Together, these results provide an input-level post hoc explanation of the routing decisions.


\subsubsection{\textbf{Semantic Characterization of Expert Composition.}}
The third layer of interpretability arises from a post hoc characterization of how learned expert compositions vary across disasters and geographic space. Figure~\ref{fig:interpretability}(e) shows distinct yet overlapping routing patterns across the four hazard types. Hurricanes, for example, assign greater weights to E1 and E4, whereas floods place more emphasis on E2. Given the associations between experts and recovery types, these patterns suggest that hurricanes more frequently combine dynamics associated with mild disruption and faster recovery, whereas floods emphasize prolonged disruption. These patterns echo previous empirical findings \cite{result-4.3-02}.

Figure~\ref{fig:interpretability}(f) maps each tract by its dominant expert, revealing substantial within-event heterogeneity. For instance, most tracts affected by the Vermont flood are dominated by E2, whereas a few activate E5, which receives the largest weight among samples outside the four types. These maps can therefore flag localized anomalies for ad hoc response. Moreover, unlike standalone post hoc analyses \cite{intro02-01, lr-05}, ReMoE has the potential to provide comparable prospective spatial characterizations across disasters.

\subsection{Robustness and Sensitivity Analysis (EQ3)}
\label{sec:uncertainty}
\begin{figure}[t]
\centering
\includegraphics[width=\columnwidth]{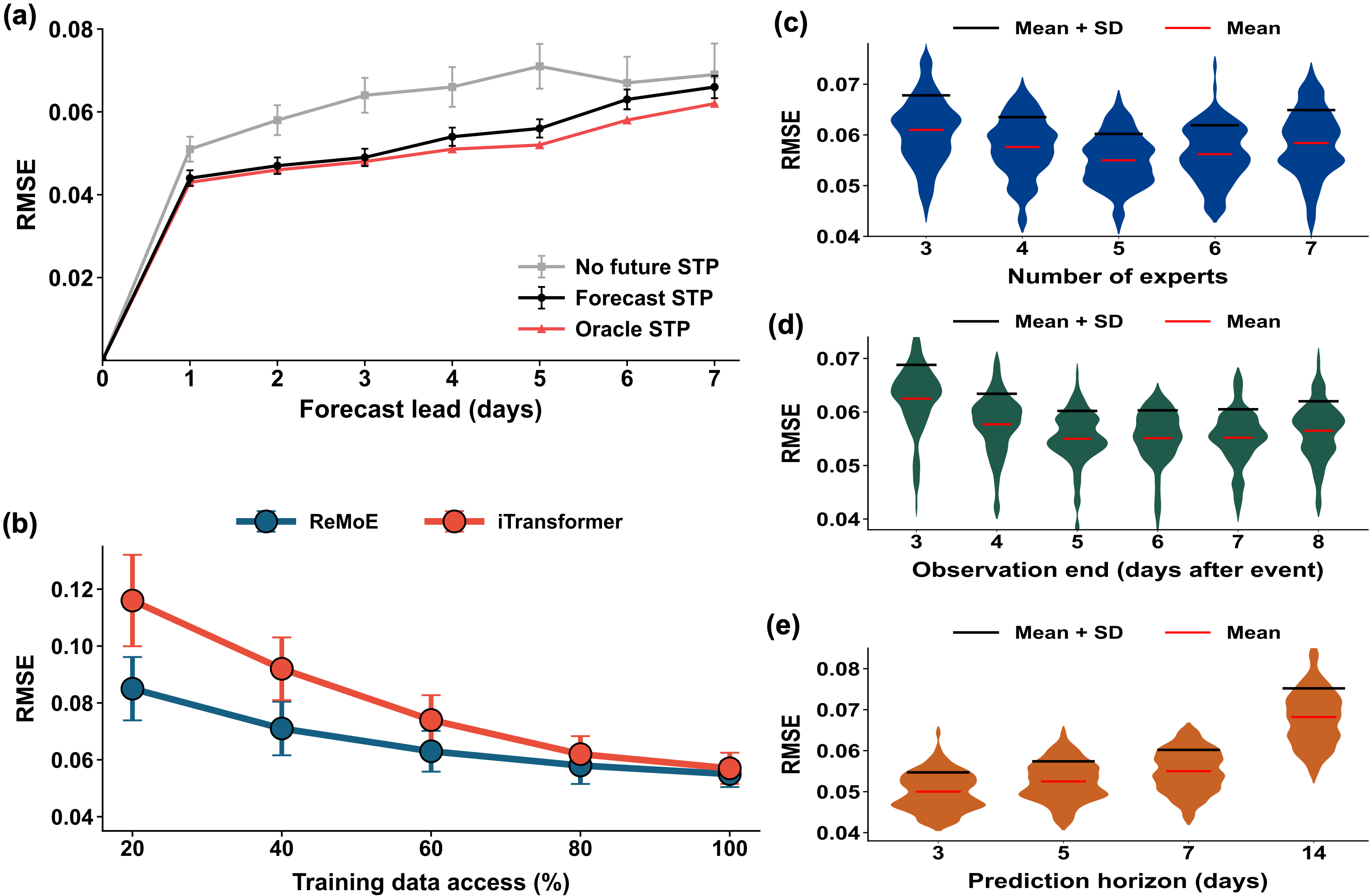}
\caption{\textbf{Robustness and sensitivity analysis.}
(a) RMSE by forecast lead under no-future, forecast, and oracle STP.
(b) ReMoE and iTransformer across training-event access levels.
(c)--(e) Sensitivity to expert count, prediction horizon, and observation endpoint.
All experiments use the ID setting.}
\Description{Five-panel robustness analysis comparing future STP conditions, training-event availability, numbers of experts, prediction horizons, and post-event observation endpoints using RMSE.}
\label{fig:uncertainty}
\end{figure}

Under realistic deployment, practical conditions may vary. We therefore evaluate ReMoE's robustness and sensitivity. First, we examine uncertainty in future hazard conditions. In Figure~\ref{fig:uncertainty}(a), Forecast STP consistently achieves lower errors than the no-future setting and approaches Oracle STP performance, highlighting the value of forecast hazard information. The increasing gap between Forecast and Oracle STP at longer leads may be due to the reduced reliability of hazard forecasts over time. Figure~\ref{fig:uncertainty}(b) evaluates ReMoE performance under limited training data. Compared with the strongest baseline under the ID setting (i.e., iTransformer), ReMoE exhibits less performance degradation as less training data are available. Its advantage is most pronounced with only \(20\%\) of the training data.

We further assess sensitivity to model configurations. In Figure~\ref{fig:uncertainty}(c), ReMoE remains stable across different numbers of experts when Top-K is fixed at 2, with five experts achieving the lowest mean RMSE and variability. We therefore adopt five experts as a balance between performance and interpretability. In addition, different applications require different prediction horizons. In practice, longer horizons are desirable if accuracy can be maintained. As shown in Figure~\ref{fig:uncertainty}(d), performance degrades moderately from 3 to 7 days but declines substantially at 14 days; thus, we select a 7-day horizon. Finally, extending the observation endpoint from \(t_0+3\) to \(t_0+5\) improves performance, whereas later observations provide limited additional gains, as shown in Figure~\ref{fig:uncertainty}(e). We therefore choose \(t_0+5\) as the earliest endpoint achieving stable accuracy.

\section{Discussion and Conclusion}
\label{sec:discussion}
\noindent\paragraph{\textbf{Potential Applications.}}
Disasters are sudden and may strike cities with limited prior experience of a particular hazard. ReMoE provides a cross-disaster prediction framework, that is to say, our predicted post-disaster mobility patterns can be used as a reference for policymakers to identify fragile regions even under unseen hazards, and decide where to assess first \cite{intro02}. This work further motivates future research on resilience curve prediction in other community systems, such as infrastructure networks \cite{discussion-01-01}.

From a broader AI-for-disaster perspective, we highlight the value of integrating scientific process knowledge with data-driven learning. Our results further suggest that model capacity per se may not be a ``silver bullet'' for complex disaster-forecasting tasks such as ours: generic deep forecasters degrade under out-of-distribution conditions, while black-box predictions remain difficult to interpret in high-stakes decision contexts \cite{intro02-02}. ReMoE addresses these challenges through a feature-guided MoE architecture whose experts are associated with distinct recovery types, providing an interpretable prediction pathway. These traceable forecasts can support anticipatory assessment, although resource decisions still require field information.


\noindent\paragraph{\textbf{Conclusion.}}
This paper introduces an interpretable mixture-of-experts framework (namely, ReMoE) to forecast human mobility recovery. ReMoE is designed to generalize to unseen disasters while providing model interpretability. To this end, we devise an IRTC module that integrates multiple specialized experts through weights learned from feature-guided sparse router to compose heterogeneous recovery types. IRTC outputs a basic recovery pattern that is forwarded into STP-RC module, enabling the prediction to adapt to evolving future hazard conditions. Extensive experiments on a large-scale multi-disaster mobility dataset demonstrate ReMoE's superior forecasting performance, particularly for unseen disasters, while interpretability analyses show transparent routing behavior and expert specialization associated with recovery types.

\section{Limitations and Ethical Considerations}
\label{sec:limitations-ethics}
Our study shares several limitations with prior research using large-scale mobility data \cite{Li2022, limitation-03}. First, the visit records cover only a subset of the population, and the data are aggregated at the census tract level, limiting analyses at smaller spatial units. Second, our 18 events were selected for clear timelines, so more disaster cases, including compound events, are needed to draw robust conclusions. Third, while we applied a consistent observation window (\(t_0-9\) to \(t_0+5\)) across all events, future research should explore more context-specific definitions of this window. Finally, associations among router features, expert weights, and empirical recovery types can only be interpreted as statistical rather than causal evidence.

\section{Generative AI Usage}
\label{sec:genai-usage}
To OpenAI Codex, for assisting with experimental implementation and language polishing.

\bibliographystyle{ACM-Reference-Format}
\bibliography{references}

\appendix
\section{Literature Review}

\begin{table*}
\centering
\caption{Existing studies on post-disaster mobility prediction, where T and ST denote temporal and spatio-temporal models, respectively, Interp. denotes model interpretability, and Cross denotes cross-disaster generalization.}
\label{tab:post-disaster-mobility-review}
\small
\begin{tabularx}{\textwidth}{
    p{1.55cm}
    p{1.55cm}
    p{0.50cm}
    p{1.85cm}
    X
    p{1.20cm}
    p{1.15cm}
    p{0.50cm}
    p{0.50cm}
}
\toprule
\textbf{Literature} &
\textbf{Target} &
\textbf{T/ST} &
\textbf{Method} &
\textbf{Features} &
\textbf{Spatial unit} &
\textbf{Temporal setting} &
\textbf{Interp.} &
\textbf{Cross} \\
\midrule

\multicolumn{9}{l}{\textbf{Statistical / probabilistic methods}} \\

\citet{s3} &
Evacuation mobility &
ST &
Markov decision process &
Historical mobility, home locations, evacuation-network
attributes, travel attributes &
10 km grid &
12--81 d &
\cmark &
\xmark \\

\citet{s2} &
Active population count &
T &
Poisson regression &
Historical mobility, temporal covariates, weather covariates,
location attributes &
900 m grid &
30--180 d &
\cmark &
\xmark \\

\citet{s4} &
Emergency behavior (Individual) &
ST &
HMM &
Historical mobility, important locations, disaster intensity, distance to hazard, government/news reports, population-flow attributes & Regional &
4 h--1 d &
\cmark &
\xmark \\

\citet{s5} &
POI visits &
T & Hawkes process &
Historical POI visits, POI categories, dwell time, temporal covariates &
Regional &
1--3 h &
\cmark &
\xmark \\


\midrule
\multicolumn{9}{l}{\textbf{Machine-learning methods}} \\

\citet{m3} &
Evacuation destination (Individual) &
ST &
XGBoost &
Mobile-phone usage, historical mobility, mobility regularity,
and mobility diversity &
Regional &
Event period &
\xmark &
\xmark \\

\citet{m1} &
Traffic flow &
T &
RF/SVM/XGBoost &
Historical traffic states, occupancy, and temporal covariates &
Road corridor &
5--20 min &
\cmark &
\xmark \\

\citet{m2} &
Evacuation OD flow &
ST &
XGBoost &
OD distance, population, and hazard attributes &
Regional &
Event period &
\xmark &
\xmark \\

\midrule
\multicolumn{9}{l}{\textbf{Deep-learning methods}} \\

\citet{m5} &
Evacuation mobility (Individual) &
ST &
DBN &
Historical mobility trajectories, hazard attributes,
transportation-network attributes, and news signals &
County &
6 h &
\cmark &
\xmark \\

\citet{m4} &
Crowd distribution &
T &
LSTM &
Historical crowd distributions and land-use attributes &
500 m grid &
Hourly &
\cmark &
\xmark \\

\citet{d1} &
Traffic speed &
T &
LSTM-NN &
Historical traffic states from upstream, target, and downstream links &
Freeway links &
5--15 min &
\xmark &
\xmark \\

\citet{d2} &
Next location (Individual) &
T &
GRU &
Recent location sequences and temporal covariates &
1 km grid &
15 min &
\xmark &
\xmark \\

\citet{d3} &
Traffic volume &
T &
LSTM &
Historical traffic states, landfall timing, social-media signals,
evacuation orders, and weather covariates &
Evacuation region &
1--30 h &
\xmark &
\xmark \\

\citet{m6} &
Crowd density/flow &
ST &
ConvLSTM &
Historical crowd-density and crowd-flow tensors &
500 m grid &
30--60 min &
\xmark &
\cmark \\

\citet{m10} &
Mobility tensor &
ST &
GCRU &
Historical multimodal mobility, temporal covariates,
and adaptive graph edges &
1 km grid &
8 steps &
\xmark &
\cmark \\

\citet{m7} &
Mobility frequency &
ST &
GCN-LSTM &
Historical mobility, social-vulnerability attributes,
built-environment, and weather covariates &
Census tract &
Daily &
\cmark &
\xmark \\

\citet{m9} &
Crowd flow &
ST &
Graph-GRU + meta-learning &
Historical mobility, social-media signals, spatial-context attributes,
and temporal covariates &
County &
6 h/6 h &
\xmark &
\cmark \\

\citet{m8} &
Mobility inflow &
ST &
GCN + Seq2Seq-LSTM &
Historical OD flows, adjacency structure,
and network-centrality attributes &
Urban grids &
6 h/1 h &
\xmark &
\xmark \\

\citet{m12} &
Mobility graph &
ST &
Graph Neural ODE &
Initial abnormal mobility graph, baseline mobility graph,
and node/edge mobility attributes &
County &
Daily &
\xmark &
\xmark \\

\citet{m11} &
Mobility flow probabilities &
ST &
Dynamic graph learning &
Pre-event mobility flows, node attributes, weather covariates,
and edge attributes &
Census tract &
Daily &
\xmark &
\xmark \\

\midrule
\multicolumn{9}{l}{\textbf{LLM-enhanced methods}} \\

\citet{m13} &
Mobility volume &
ST &
ST-ResNet/GCN + LLM &
Historical mobility, POI attributes, temporal/region embeddings,
OD graph, and LLM-derived event-intention features &
Regional &
1--24 h &
\xmark &
\cmark \\

\citet{m14} &
Next location (Individual) &
ST &
LLM-enhanced hybrid model &
Cross-city mobility trajectories, hazard attributes,
and LLM-derived intention features &
Candidate locations &
Next step &
\xmark &
\xmark \\

\bottomrule
\end{tabularx}

\medskip
\footnotesize
\textit{Abbreviations:}
Markov decision process (MDP), hidden Markov model (HMM), random forest (RF),
support vector machine (SVM), extreme gradient boosting (XGBoost),
deep belief network (DBN), long short-term memory (LSTM),
gated recurrent unit (GRU), convolutional LSTM (ConvLSTM),
graph convolutional network (GCN),
graph convolutional recurrent unit (GCRU),
neural ordinary differential equation (Neural ODE),
spatio-temporal residual network (ST-ResNet).
\end{table*}


\subsection{Post-Disaster Human Mobility Forecasting}
\label{sec:post-disaster-mobility-forecasting}

By post-disaster human mobility, we refer to the short-term disruption and recovery of human movement following a disaster. Early studies mainly adopted probabilistic and statistical approaches. \citet{s3} modeled evacuation behavior as a Markov decision process based on an evacuation graph, while \citet{s4} used a hidden Markov model. Despite their interpretability, these methods relied on simplifying assumptions and were largely event-specific. Machine learning (ML) methods subsequently introduced models such as XGBoost to predict evacuation destinations and origin--destination flows using historical mobility, demographic, and hazard-exposure features \cite{m2,m3}. However, these approaches remained limited in capturing spatio-temporal dependencies.

DL methods were subsequently introduced to learn nonlinear mobility dynamics. Early DL approaches mainly adopted recurrent architectures. For instance, \citet{m4} introduced an LSTM-based model to learn temporal mobility patterns from sequential observations. However, mobility dynamics are inherently spatio-temporal, motivating models that jointly capture temporal evolution and spatial interactions. Architectures such as ConvLSTM and GCN--LSTM combined recurrent networks with convolutional or graph-based operators to learn regional dependencies and dynamic mobility patterns \cite{m6,m7}. Recent studies have incorporated adaptive graph learning and
LLM-derived external signals \cite{m9, m13}, while Neural ODEs model mobility recovery as a continuous dynamic process \cite{m12}. However, these models remain difficult to interpret, and their spatial representations may overfit location-specific dependencies, limiting transferability to unseen regions.

\subsection{Interpretability in Post-disaster Mobility Forecasting}
Interpretability is essential for post-disaster mobility forecasting, as transparent predictions can help policymakers understand mobility disruptions and design effective recovery strategies. However, it remains underexplored in existing post-disaster mobility forecasting studies. Traditional statistical and ML models provide relatively direct interpretations through explicit coefficients, decision structures, or feature importance analyses \cite{s1,s2,s3,m1,m4,m5,m7}. In contrast, DL models are more difficult to interpret due to their nonlinear architectures, with only limited efforts exploring post-hoc explanations. For example, \citet{m4} examined temporal and spatial correlations to characterize learned mobility patterns, while \citet{m7} applied DeepLIFT-based attribution to a GCN--LSTM model to quantify the contribution of community attributes to predicted mobility disruptions. However, existing studies rarely go beyond post-hoc, feature-level explanations, leaving the intrinsic interpretability of forecasting models largely unexplored \cite{method-3.2.2-yang, lyu2025srl, Lyu2025conformancechecking}. ReMoE bridges this gap by combining both: its feature-guided router makes expert allocation traceable, while post-hoc analysis connects learned expert specialization with empirical recovery types, revealing the distinct semantic roles of different experts.


\subsection{Cross-Disaster Generalization}
\label{app:literature_review_cross_disaster}
To support effective disaster management, mobility forecasting models must be capable of predicting responses to unprecedented events \cite{LYU2026105787}. However, mobility responses vary across locations, hazard types, and severities, making models trained on specific cases difficult to generalize to unseen disasters. Celebrated approaches have been explored to address this challenge, including transfer learning strategies that transfer mobility patterns across locations and reinforcement learning methods that represent disaster impacts through inferred reward functions \cite{d2, d1}. Although these approaches are promising contributions, limitations remain in their generalizability. For example, existing methods improve transferability by transferring observed patterns or learned representations rather than the underlying recovery dynamics shared across disasters. Inspired by prior work on modeling cross-disaster recovery dynamics \cite{Li2022, intro02, intro03-01, lr-05}, ReMoE addresses this gap by learning reusable recovery types through expert specialization and enabling transferable prediction.

\section{Dataset and Recovery Types}
\label{app:dataset and recovery types}

\subsection{Dataset Description}
\label{app:dataset_description}

\begin{figure*}[t]\centering\includegraphics[width=\textwidth]{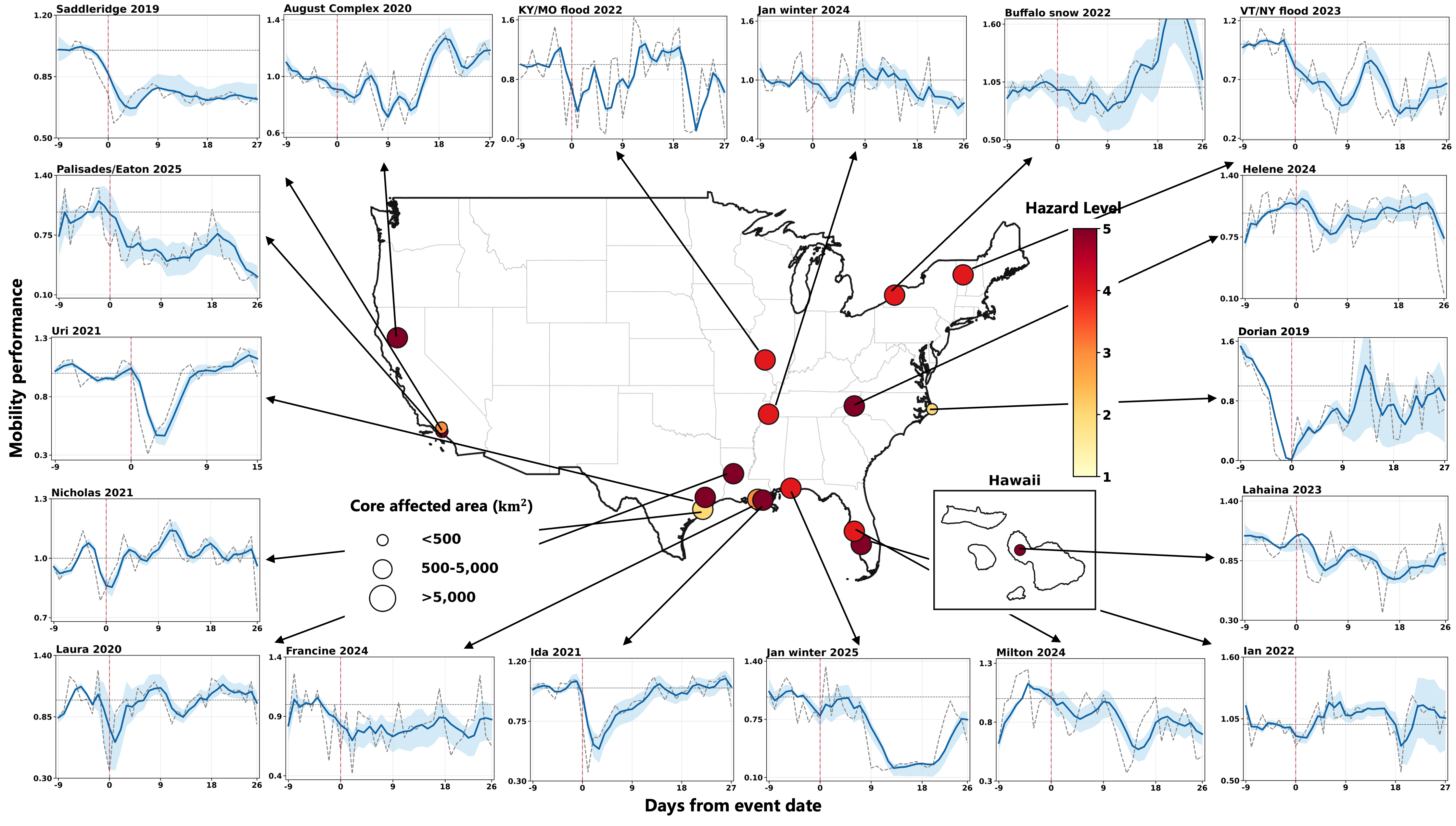}\caption{\textbf{Disaster events and mobility recovery curves.} Event locations use marker size and color to indicate affected-area extent and hazard severity. Panels show normalized mobility: blue lines and shading
indicate the mean and interquartile range of three-day causally smoothed tract
curves; gray dashed lines indicate the unsmoothed mean. Red lines mark the
event date.}\Description{A U.S. map showing 18 hurricanes, floods, winter storms,and wildfires, surrounded by their normalized tract-level mobilityrecovery curves.}\label{fig: study_areas}\end{figure*}

\subsubsection{\textbf{Event Selection.}}
\label{app: event selection}
Our multi-disaster dataset spans different severity levels. As shown in Fig.~\ref{fig: study_areas}, it contains 18 major U.S. disasters between 2019 and 2025, including eight hurricanes, four winter storms,
two floods, and four wildfires. Events were selected when (i) their chronology and principal impact area could be established from official records, (ii) Veraset provided sufficient pre- and post-event mobility coverage, and (iii) interference from concurrent or consecutive disasters could be minimized. The reference event date \(t_0\), study area, and hazard severity of each event were determined from official hazard-specific reports and classifications. Table~\ref{tab:event_inventory_18} summarizes each event's metadata, analysis window, and data sources. All future weather come from a single forecast issued at the forecast origin ($t_0+5$). Beyond a product's forecast range, we keep its last value unchanged. For example, NWM flood forecasts cover 240~h, so in the 14-day test, days 11--14 reuse the day-10 value.

\subsubsection{\textbf{Mobility Processing.}}
For each event, raw visit records were mapped to U.S. Census tracts and ordered by device and timestamp. Consecutive visits to different tracts were converted into directed movements and aggregated by day to obtain
the daily inter-tract flows \(f_{ij,e}(\tau)\). Following
Definition~2.1, the total mobility of each tract was calculated by summing its daily incoming and outgoing inter-tract movements. We then estimated tract-specific mobility baselines using the two complete weeks preceding the observation window, \([\tau_0^e-23,\tau_0^e-10]\). Separate weekday and weekend baselines were used to account for calendar effects. Tracts with fewer than
50 mean daily inter-tract movements during this baseline period were excluded to reduce sampling bias in sparsely observed areas. The remaining daily mobility values were normalized by their corresponding weekday or weekend baselines and subsequently processed using a three-day causal moving average.

\subsubsection{\textbf{Temporal Window Design.}}
\label{app:temporal window design}
For each event, data were aligned to the reference date \(t_0\) and collected from \(t_0-23\) to \(t_0+19\). The interval \([t_0-23,t_0-10]\) forms the 14-day baseline period, \([t_0-9,t_0+5]\) the 15-day observation window, and
\([t_0+6,t_0+19]\) the maximum prediction horizon used in sensitivity analysis. The primary seven-day horizon spans \([t_0+6,t_0+12]\). The observation window captures the maximum impact and early recovery signals used to construct the router features \cite{intro03-01}. Early-recovery is defined at the tract level as the first rebound of at least
\(10\%\) of the maximum impact depth toward baseline. A post-hoc check confirmed that over \(95\%\) of the samples met the required criteria. The seven-day primary prediction horizon is motivated by two considerations: reported recovery intervals generally span 18--168 hours \cite{experiment4.1-dataset-01}, while horizons over a week extend beyond the reliable weather forecast window and may introduce additional noise into the model.



\begin{table*}
\centering
\small
\setlength{\tabcolsep}{2.0pt}
\renewcommand{\arraystretch}{1.10}

\newcommand{\eventhdr}[1]{%
  \begin{minipage}[c][3.2em][c]{\linewidth}%
    \centering\bfseries #1%
  \end{minipage}%
}

\caption{Event inventory, common data-collection windows, and hazard-data
sources for the 18 disaster events.}
\label{tab:event_inventory_18}

\begin{tabularx}{\textwidth}{
@{}
>{\raggedright\arraybackslash}p{3.15cm}
>{\centering\arraybackslash}p{1.15cm}
>{\centering\arraybackslash}p{1.45cm}
>{\centering\arraybackslash}p{3.05cm}
>{\centering\arraybackslash}p{1.35cm}
>{\centering\arraybackslash}p{2.10cm}
>{\centering\arraybackslash}X
@{}
}

\toprule
\eventhdr{Event}
& \eventhdr{Affected\\tracts}
& \eventhdr{Event start\\date ($t_0$)}
& \eventhdr{Data collection\\time frame}
& \eventhdr{Affected\\users}
& \eventhdr{Disaster\\intensity}
& \eventhdr{Data sources and roles} \\
\midrule

\multicolumn{7}{l}{\textbf{Hurricanes}} \\[2pt]

Dorian (2019)
& 24
& 2019/9/6
& 2019/8/14--2019/9/25
& 60,341
& \multirow[c]{8}{2.10cm}{\centering Rainfall}
& \multirow[c]{8}{=}{%
  \parbox{\linewidth}{\centering
    Hurricane reports: NHC \cite{appendix-b1-nhc-tcr}\par
    Hurricane tracks: HURDAT2 \cite{appendix-b2-hurdat2}\par
    Official affected areas: FEMA \cite{appendix-b3-openfema}\par
    Past weather: NOAA PSL \cite{appendix-b4-noaa-psl}\par
    Forecast weather: GFS \cite{appendix-b5-gfs}
  }
} \\

Laura (2020)
& 464
& 2020/8/27
& 2020/8/4--2020/9/15
& 510,594
& & \\

Ida (2021)
& 403
& 2021/8/29
& 2021/8/6--2021/9/17
& 319,843
& & \\

Nicholas (2021)
& 127
& 2021/9/14
& 2021/8/22--2021/10/3
& 218,065
& & \\

Ian (2022)
& 341
& 2022/9/28
& 2022/9/5--2022/10/17
& 376,873
& & \\

Francine (2024)
& 398
& 2024/9/11
& 2024/8/19--2024/9/30
& 1,171,640
& & \\

Helene (2024)
& 211
& 2024/9/27
& 2024/9/4--2024/10/16
& 635,960
& & \\

Milton (2024)
& 491
& 2024/10/9
& 2024/9/16--2024/10/28
& 1,929,541
& & \\

\midrule

\multicolumn{7}{l}{\textbf{Winter storms}} \\[2pt]

Uri (2021)
& 786
& 2021/2/14
& 2021/1/22--2021/3/5
& 1,080,699
& \multirow[c]{4}{2.10cm}{\centering Temperature drop}
& \multirow[c]{4}{=}{%
  \parbox{\linewidth}{\centering
    Winter-storm reports: NWS \cite{appendix-b6-nws}\par
    Official affected areas: FEMA \cite{appendix-b3-openfema}\par
    Past weather: GHCN \cite{appendix-b7-ghcn-daily}\par
    Forecast weather: GFS \cite{appendix-b5-gfs}
  }
} \\

Buffalo lake-effect snowstorm (2022)
& 185
& 2022/11/18
& 2022/10/26--2022/12/7
& 271,682
& & \\

January winter storm (2024)
& 374
& 2024/1/15
& 2023/12/23--2024/2/3
& 2,705,099
& & \\

January winter storm (2025)
& 543
& 2025/1/21
& 2024/12/29--2025/2/9
& 2,068,469
& & \\

\midrule

\multicolumn{7}{l}{\textbf{Floods}} \\[2pt]

KY--MO riverine flood (2022)
& 391
& 2022/7/28
& 2022/7/5--2022/8/16
& 622,394
& \multirow[c]{2}{2.10cm}{\centering Streamflow}
& \multirow[c]{2}{=}{%
  \parbox{\linewidth}{\centering
    Flood reports: NWS \cite{appendix-b6-nws}\par
    Official affected areas: FEMA \cite{appendix-b3-openfema}\par
    Past weather: USGS NWIS \cite{appendix-b8-usgs-nwis}\par
    Forecast weather: NWM \cite{appendix-b9-nwm}
  }
} \\

Vermont--New York flood (2023)
& 108
& 2023/7/10
& 2023/6/17--2023/7/29
& 40,419
& & \\

\midrule

\multicolumn{7}{l}{\textbf{Wildfires}} \\[2pt]

Saddleridge (2019)
& 27
& 2019/10/10
& 2019/9/17--2019/10/29
& 119,149
& \multirow[c]{4}{2.10cm}{\centering Near-Surface Smoke}
& \multirow[c]{4}{=}{%
  \parbox{\linewidth}{\centering
    Wildfire reports: NIFC \cite{appendix-b10-nifc-perimeters}\par
    Fire perimeters: NIFC \cite{appendix-b10-nifc-perimeters}\par
    Past weather: HRRR-Smoke \cite{appendix-b12-hrrr}\par
    Forecast weather: HRRR-Smoke \cite{appendix-b12-hrrr}
  }
} \\

August Complex (2020)
& 88
& 2020/8/16
& 2020/7/24--2020/9/4
& 72,919
& & \\

Lahaina (2023)
& 45
& 2023/8/8
& 2023/7/16--2023/8/27
& 29,646
& & \\

Palisades--Eaton (2025)
& 25
& 2025/1/7
& 2024/12/15--2025/1/26
& 117,850
& & \\

\bottomrule
\end{tabularx}

\vspace{2pt}

\begin{minipage}{\textwidth}
\footnotesize
\textit{Note.}
Because the datasets vary by event, only the source platforms are listed here. Detailed event-level sources are available in our repository.
\end{minipage}

\end{table*}

\subsection{Post-hoc Recovery Type Identification}
\label{app:post-hoc type}

\subsubsection{\textbf{Clustering the recovery curves.}}

Unlike studies that cluster complete recovery processes
\cite{intro03-01,appendix-b2-01}, we use a fixed 22-day window,
\([\tau_0^e-9,\tau_0^e+12]\), to ensure consistent temporal coverage
across events. Using the mobility curves processed as described above,
the clustering input for each tract--event sample is
\begin{equation}
\mathbf{x}_{i,e}
=
\left[
\widetilde m^{\mathrm{sm}}_{i,e}(\tau_0^e-9),
\ldots,
\widetilde m^{\mathrm{sm}}_{i,e}(\tau_0^e+12)
\right]^{\top}
\in\mathbb{R}^{22}.
\end{equation}
We use DTW K-means because modest differences in disruption and recovery
timing may cause otherwise similar curves to be misclassified under
pointwise Euclidean distance. For any two clustering inputs
$\mathbf{x}$ and
$\mathbf{y}$, their DTW distance is
\begin{equation}
d_{\mathrm{DTW}}(\mathbf{x},\mathbf{y})
=
\min_{\pi}
\left[
\sum_{(u,v)\in\pi}(x_u-y_v)^2
\right]^{1/2},
\end{equation}
where $u$ and $v$ index days in $\mathbf{x}$ and $\mathbf{y}$, respectively,
and $\pi$ is a valid warping path constrained by $|u-v|\leq2$. This allows small timing differences without aligning physically distinct recovery dynamics. For candidate cluster number $K$, the within-cluster distortion is
\begin{equation}
D_K
=
\sum_{k=1}^{K}
\sum_{\mathbf{x}_{i,e}\in C_k}
d_{\mathrm{DTW}}
\left(\mathbf{x}_{i,e},\boldsymbol{\mu}_k\right)^2,
\end{equation}
where $C_k$ and $\boldsymbol{\mu}_k$ are cluster $k$ and its DTW barycenter.
We use K-means++ initialization and multiple random restarts. Then, we evaluate $K=2,\ldots,8$ using the DTW-based silhouette coefficient
\begin{equation}
s(i)
=
\frac{b(i)-a(i)}
{\max\{a(i),b(i)\}},
\end{equation}
where $a(i)$ is the mean within-cluster distance and $b(i)$ is the minimum mean distance to another cluster. We selected $K=4$ considering the average silhouette score, the elbow in $D_K$ and requiring that each cluster span multiple events rather than be dominated by a single large event. The clustering is stable. The silhouette score is highest at $K=4$ (0.34 vs.\ 0.23--0.31), and the clusters change little across 10 random seeds, 100 event-stratified bootstrap resamples, and 18 leave-one-event-out runs (mean adjusted Rand index of 0.90, 0.87, and 0.83, respectively), with all four recovery types preserved. The choice of $K$ is independent of the number of ReMoE experts. 

\subsubsection{\textbf{Clustered recovery types.}}
After selecting $K=4$, we represent each cluster by the pointwise mean of its event-aligned curves. As shown in Figure~\ref{fig:interpretability}(a), the four recovery types are:

\begin{itemize}
    \item \textbf{Type I: Rapid rebound.} A mild mobility disruption followed by a rapid V-shaped return to baseline, commonly associated with short-duration hurricanes or winter storms where mobility resumes quickly after the hazard passes.
    
    \item \textbf{Type II: Long deficit.} A pronounced disruption followed by persistently low mobility and limited recovery, typically observed in severely affected tracts following destructive floods, or wildfires where infrastructure damage, service interruption, or displacement is prolonged. \cite{lr-05}.
    
    \item \textbf{Type III: Staged recovery.} An initial rapid rebound followed by a clear reduction in recovery rate. This pattern often arises after hurricanes or floods when emergency access is restored quickly but broader infrastructure, businesses, and services reopen in stages \cite{intro02-01}.
    
    \item \textbf{Type IV: Slow monotonic recovery.} A slow, approximately monotonic return toward baseline with a low
    initial rebound, typically observed in severely affected tracts with substantial functional loss. \cite{intro03-01}.
\end{itemize}


\section{Router Feature Construction}
\label{app:router-features}

\begin{table*}
\centering
\footnotesize
\setlength{\tabcolsep}{3.0pt}
\renewcommand{\arraystretch}{1.10}
\renewcommand{\multirowsetup}{\centering}
\caption{Definitions and sources of the 16 router features used by ReMoE.}
\label{tab:app-router-features}

\begin{tabularx}{\textwidth}{
@{}
>{\centering\arraybackslash}m{1.65cm}
>{\centering\arraybackslash}m{1.35cm}
>{\centering\arraybackslash}m{1.70cm}
>{\raggedright\arraybackslash}m{5.85cm}
>{\raggedright\arraybackslash}X
@{}
}
\toprule
\textbf{Category}
& \textbf{Type}
& \textbf{Metric}
& \textbf{Formula/Source}
& \textbf{Definition} \\
\midrule


\multirow[c]{5}{=}[-8.5ex]{\textbf{Impact}}
& Magnitude
& DI
& $\displaystyle
  \mathrm{DI}_{i,e}=
  \left[1-{\mathbf m}^{\mathrm{in},\min}_{i,e}\right]_{+}$
& Maximum post-event mobility loss within the early observation window. \\
\cmidrule(lr){2-5}

& Duration
& Timp
& $\displaystyle
  \mathrm{Timp}_{i,e}=
  \frac{\tau^{\min}_{i,e}-\tau_0^e}
       {\tau_{\mathrm{in}}^e-\tau_0^e}$
& Normalized time required to reach the maximum observed impact. \\
\cmidrule(lr){2-5}

& Integral
& ImpPersist
& $\displaystyle
  \mathrm{ImpPersist}_{i,e}=
  \frac{
  \int_{\tau_0^e}^{\tau^{\min}_{i,e}}
  [1-{\mathbf m}^{\mathrm{in}}_{i,e}(\tau)]_{+}\,d\tau}
  {\mathrm{DI}_{i,e}
  (\tau^{\min}_{i,e}-\tau_0^e)}$
& Normalized cumulative impact before the observed minimum; larger values
indicate a more persistent impact phase. \\
\cmidrule(lr){2-5}

& Rate
& RelFR
& $\displaystyle
  \mathrm{RelFR}_{i,e}=
  \frac{
  [{\mathbf m}^{\mathrm{in}}_{i,e}(\tau_0^e)
  -{\mathbf m}^{\mathrm{in},\min}_{i,e}]_{+}}
  {\max(\mathrm{DI}_{i,e},\epsilon)
   \max(\mathrm{Timp}_{i,e},0.2)}$
& Relative speed at which the observed mobility loss is realized. \\
\cmidrule(lr){2-5}

& Censoring
& Censored
& $\displaystyle
  \mathrm{Censored}_{i,e}=
  \mathbb{I}
  \!\left(\tau^{\min}_{i,e}=\tau_{\mathrm{in}}^e\right)$
& Indicates that the observed minimum occurs at the observation boundary. \\

\midrule


\multirow[c]{6}{=}[-10.5ex]{\textbf{Recovery}}
& \multirow[c]{2}{1.35cm}[-3.5ex]{\centering\arraybackslash Rate}
& RR
& $\displaystyle
  \mathrm{RR}_{i,e}=
  \frac{
  {\mathbf m}^{\mathrm{in}}_{i,e}(\tau_{\mathrm{in}}^e)
  -{\mathbf m}^{\mathrm{in},\min}_{i,e}}
  {\max(\tau_{\mathrm{in}}^e-\tau^{\min}_{i,e},1)}$
& Mean recovery rate from the observed minimum to the observation endpoint. \\
\cmidrule(lr){3-5}

&
& Momentum
& $\displaystyle
  {\mathbf m}^{\mathrm{in}}_{i,e}(\tau_0^e+d)=
  \alpha_{i,e}+\beta_{i,e}d+\varepsilon_{i,e,d},
  \quad d\in\{2,3,4,5\};
  \quad \mathrm{Momentum}_{i,e}=\widehat\beta_{i,e}$
& Recent recovery slope near the observation endpoint. \\
\cmidrule(lr){2-5}

& Ratio
& RRatio
& $\displaystyle
  \mathrm{RRatio}_{i,e}=
  \frac{
  {\mathbf m}^{\mathrm{in}}_{i,e}(\tau_{\mathrm{in}}^e)
  -{\mathbf m}^{\mathrm{in},\min}_{i,e}}
  {\max(\mathrm{DI}_{i,e},\epsilon)}$
& Fraction of the maximum observed mobility loss restored by the observation
endpoint. \\
\cmidrule(lr){2-5}

& \multirow[c]{2}{1.35cm}[-3.5ex]{\centering\arraybackslash Integral}
& RecDefRatio
& $\displaystyle
  \mathrm{RecDefRatio}_{i,e}=
  \frac{
  \int_{\tau^{\min}_{i,e}}^{\tau_{\mathrm{in}}^e}
  [1-{\mathbf m}^{\mathrm{in}}_{i,e}(\tau)]_{+}\,d\tau}
  {\max(\mathrm{DI}_{i,e},\epsilon)
   \max(\tau_{\mathrm{in}}^e-\tau^{\min}_{i,e},1)}$
& Remaining below-baseline recovery burden normalized by impact depth and
post-minimum duration. \\
\cmidrule(lr){3-5}

&
& AB
& $\displaystyle
  \mathrm{AB}_{i,e}=
  \int_{\tau^{\min}_{i,e}}^{\tau_{\mathrm{in}}^e}
  [{\mathbf m}^{\mathrm{in}}_{i,e}(\tau)-1]_{+}\,d\tau$
& Cumulative above-baseline mobility during early recovery. \\
\cmidrule(lr){2-5}

& Stability
& PreVar
& $\displaystyle
  \mathrm{PreVar}_{i,e}=
  \operatorname{SD}
  \!\left\{
  {\mathbf m}^{\mathrm{in}}_{i,e}(\tau_0^e-9),\ldots,
  {\mathbf m}^{\mathrm{in}}_{i,e}(\tau_0^e-1)
  \right\}$
& Pre-event variability of normalized mobility. \\

\midrule


\multirow[c]{5}{=}[-4.5ex]{\textbf{Contextual}}
& Pressure
& $\mathrm{STP}^{\mathrm{obs}}$
& NOAA \cite{lr-fl-03} (see Eq.~\eqref{eq:stp} for details)
& Mean observed spatiotemporal pressure from the mobility minimum to the
observation endpoint. \\
\cmidrule(lr){2-5}

& Vulnerability
& SVI
& CDC/ATSDR \cite{lr-fl-02}
& Tract-level social vulnerability. \\
\cmidrule(lr){2-5}

& Exposure
& PopDens
& USEPA \cite{lr-fl-01}
& Log-transformed tract population density. \\
\cmidrule(lr){2-5}

& Land use
& BuiltUp
& USEPA \cite{lr-fl-01}
& Share of developed or built-up land within the tract. \\
\cmidrule(lr){2-5}

& Network centrality
& PRbase
& Eq.~\eqref{eq:prbase}
& Event-level percentile of baseline weighted PageRank. \\

\bottomrule
\end{tabularx}

\vspace{2pt}
\begin{minipage}{\textwidth}
\footnotesize
\textit{Notation.}
We define $\tau^{\min}_{i,e}$ as the earliest day when
$\widetilde m^{\mathrm{in}}_{i,e}$ reaches its minimum over
$[\tau_0^e,\tau_{\mathrm{in}}^e]$, denote the corresponding value by
$\widetilde m^{\mathrm{in},\min}_{i,e}$, and use
$[x]_+=\max(x,0)$ with $\epsilon=10^{-4}$.
\end{minipage}
\end{table*}

Table~\ref{tab:app-router-features} details the 16 router features. Impact and recovery features are
derived from the predefined 15-day observation window
\([\tau_0^e-9,\tau_0^e+5]\). To avoid redundancy, we retain impact depth but exclude residual mobility at maximum impact, as one is mathematically determined by the other. Similarly, normalized impact timing and relative decline rate are used instead of their absolute counterparts to improve cross-event comparability. For contextual features, we average the daily STP scores over the observation window. Region-related features, including social vulnerability, population density, and built-environment characteristics, are obtained from the datasets listed in Table~\ref{tab:app-router-features}. Additionally, we include baseline network centrality to capture differences in regions' pre-event network positions. Specifically, using the mobility flows over the two weeks preceding the observation window, we construct
\begin{equation}
w^{\mathrm{base}}_{ji,e}
=
\frac{1}{14}
\sum_{\tau=\tau_0^e-23}^{\tau_0^e-10}
f_{ji,e}(\tau),
\qquad j\neq i.
\end{equation}
The baseline PageRank percentile is then
\begin{equation}
\label{eq:prbase}
\mathrm{PRbase}_{i,e}
=
\operatorname{PctRank}_{e}
\left[
\operatorname{PageRank}_{0.85}
\left(\mathbf W^{\mathrm{base}}_e\right)_i
\right].
\end{equation}
A higher $\mathrm{PRbase}_{i,e}$ indicates a more central region.

\begin{figure}[!htbp]
\centering
\includegraphics[width=\columnwidth]{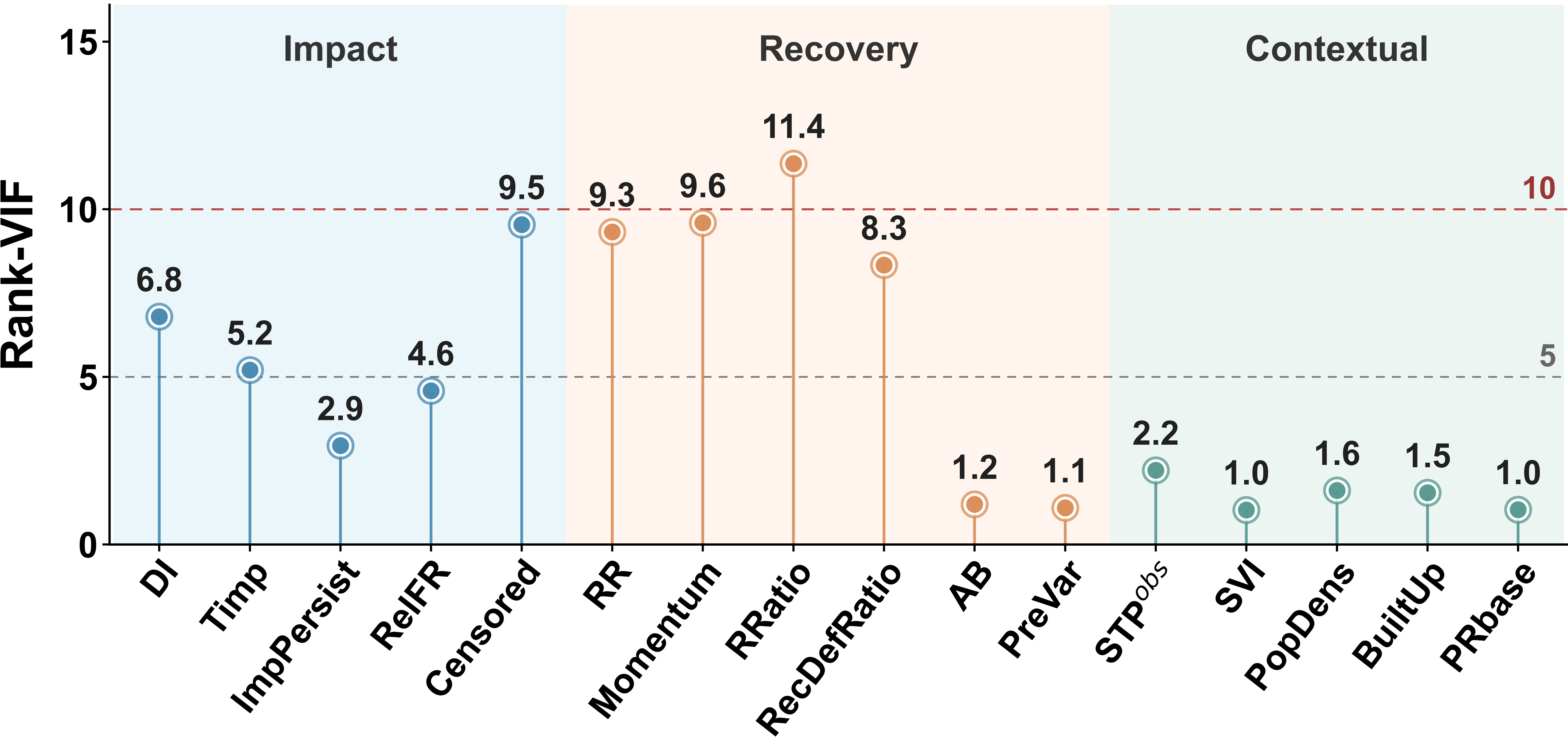}
\caption{\textbf{Rank-VIF diagnostics for the 16 router features.}}
\Description{Rank-based variance inflation factors for the 16 router
features, grouped into impact, recovery, and contextual categories.}
\label{fig:app-router-feature-filtering}
\end{figure}

Finally, we assessed feature redundancy using Rank-VIF, which captures redundancy arising from monotonic relationships. As shown in Figure~\ref{fig:app-router-feature-filtering}, most Rank-VIF values were below 10; RRatio was retained despite a slight exceedance because it captures distinct recovery progress.

\section{Additional Experiments}
\label{app:additional-results}

\begin{itemize}[leftmargin=*, itemsep=2pt, topsep=2pt]
    \item \textbf{EQ-A1:} How do different feature groups affect ReMoE's forecasting performance?
    \(\rightarrow\) Sec.~\ref{app:router-feature-ablation}
    \item \textbf{EQ-A2:} Does ReMoE learn a routing space better aligned with recovery semantics than a standard MoE?
    \(\rightarrow\) Sec.~\ref{app:routing-space-comparison}
    \item \textbf{EQ-A3:} How do input features contribute nonlinearly to ReMoE's expert assignment?
    \(\rightarrow\) Sec.~\ref{app:shap-router-attribution}
\end{itemize}

\subsection{Impact of Feature Group on Model Performance (EQ-A1)}
\label{app:router-feature-ablation}

\begin{figure}[!ht]
    \centering
    \includegraphics[width=\columnwidth]{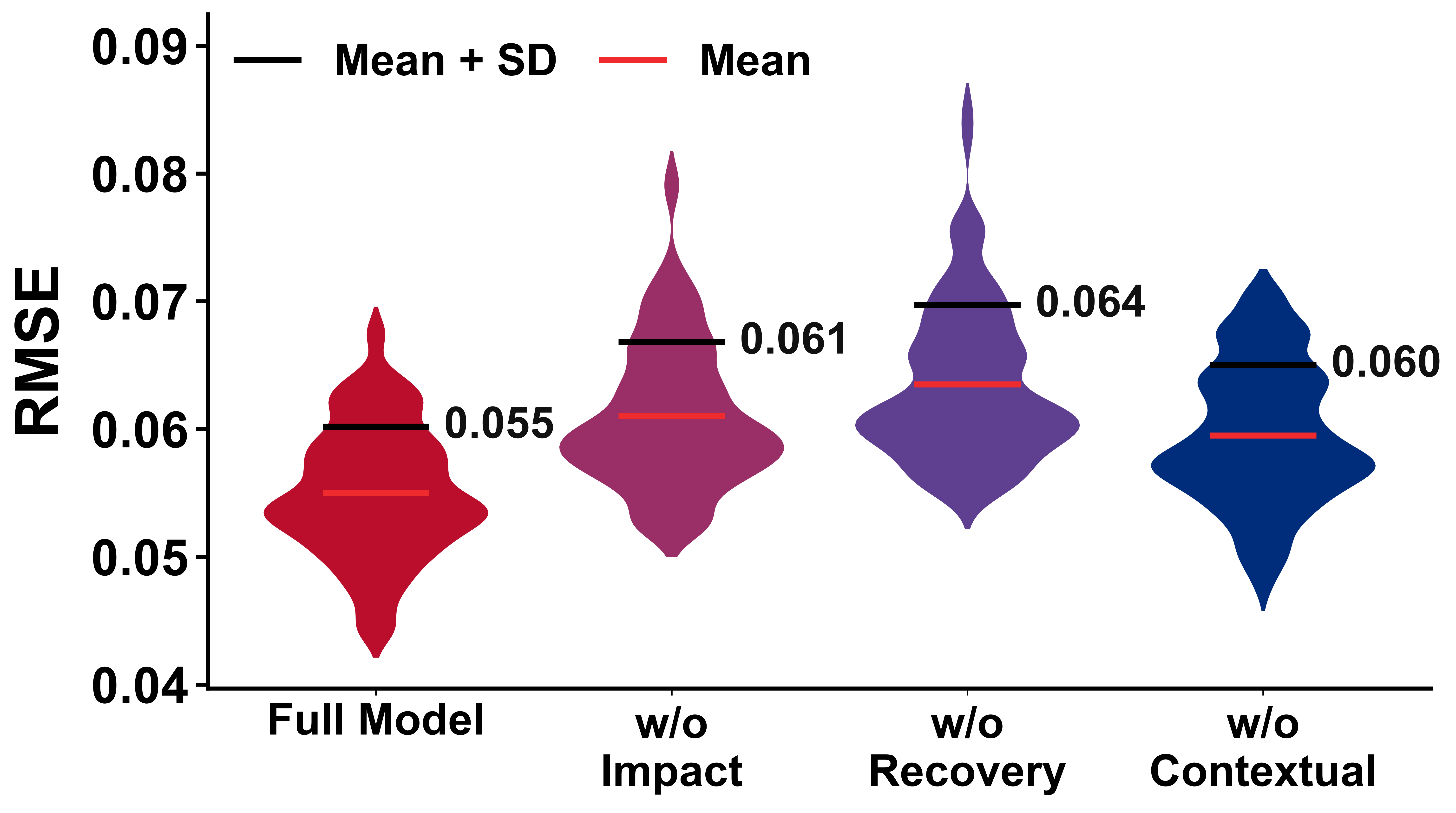}
    \caption{\textbf{Router feature-group ablation.} Tract-level RMSE distributions after neutralizing each router feature group.}
    \Description{Violin plots showing tract-level RMSE distributions for the
    full model and models without impact, recovery, or contextual router
    features.}
    \label{fig:app-router-feature-group-ablation}
\end{figure}

Here we present a feature-level supplementary ablation to evaluate the value of features groups used by router. ``Full'' denotes the complete model using all 16 router features, while ``w/o Impact,'' ``w/o Recovery,'' and ``w/o Contextual'' denote variants with the corresponding impact, recovery, or contextual feature group removed, with all other inputs unchanged.

Figure~\ref{fig:app-router-feature-group-ablation} shows the RMSE distributions on the ID test set, where red lines indicate the mean and black lines indicate the mean plus one standard deviation. The full model achieves a mean RMSE of 0.055. Removing impact features increases the mean RMSE to 0.061 (+10.9\%), removing contextual features to 0.060 (+9.1\%), and removing recovery features produces the largest degradation, reaching 0.064 (+16.4\%). This ordering suggests that recovery features may carry the largest aggregate attribution, followed by impact and contextual features, which provide complementary but smaller contributions to expert routing and forecasting. Nevertheless, retaining all feature groups yields the best overall performance.

\subsection{Routing-Space Comparison (EQ-A2)}
\label{app:routing-space-comparison}

\begin{figure}[!ht]
    \centering
    \includegraphics[width=\columnwidth]{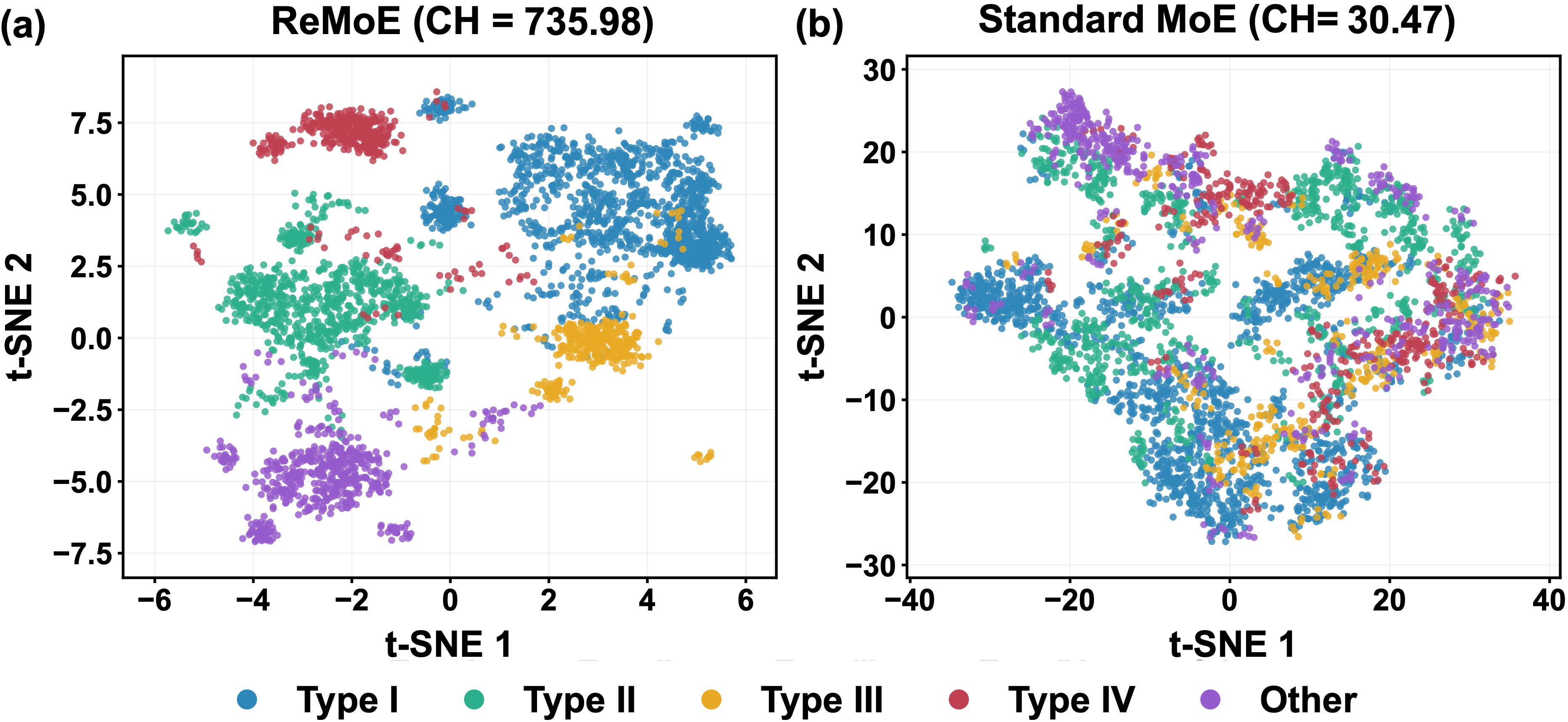}
    \caption{\textbf{Routing-space comparison.}
    t-SNE visualization of the routing spaces learned by
    ReMoE (a) and a standard MoE (b), with samples colored by post hoc
    recovery type.}
    \label{fig:app-routing-space}
\end{figure}

We examine whether the integrated MoE design of ReMoE learns a more semantically structured routing space than a standard MoE. The standard MoE uses a latent-representation-based router and conventional MLP experts. For both models, we project the pre-Top-\(K\) router logits using t-SNE and color each sample by its post hoc recovery type. As visualized in Figure~\ref{fig:app-routing-space}, ReMoE forms more coherent regions aligned with different recovery types, whereas the standard MoE exhibits greater mixing. This suggests that ReMoE not only makes expert selection easier to trace, but also yields a routing space more closely aligned with post hoc recovery types.

We quantify this effect using the Calinski–Harabasz (CH) index, which measures cluster compactness and separation. CH is computed on the original five-dimensional router logits for both models. Higher CH values indicate better-defined clusters. ReMoE achieves a substantially higher CH score than the standard MoE (735.98 vs.\ 30.47), supporting a more structured routing space.

\subsection{Nonlinear Feature Effects on Expert Routing (EQ-A3)}
\label{app:shap-router-attribution}

\begin{figure}[!ht]
    \centering
    \includegraphics[width=\columnwidth]{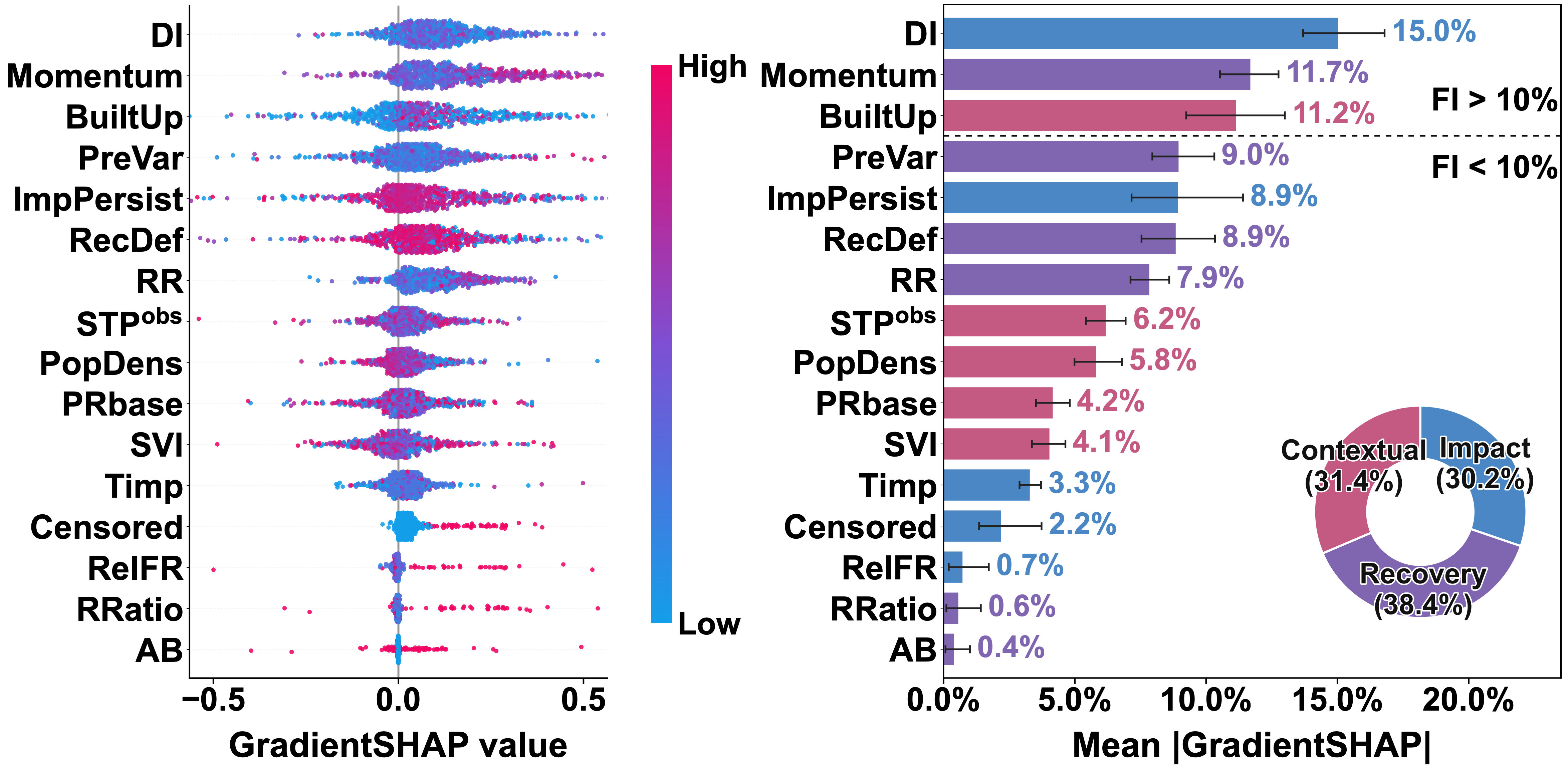}
    \caption{\textbf{Global router attribution of the 16 input features.}}
    \label{fig:app-router-attribution-global}
\end{figure}

To complement the monotonic associations in
Fig.~\ref{fig:interpretability}d, we apply GradientSHAP to quantify the nonlinear
contributions of the 16 input features to each sample's dominant expert, using its pre-Top-$K$ softmax probability as the target. The left beeswarm in Figure~\ref{fig:app-router-attribution-global} shows signed attributions to each sample's dominant expert, with color indicating feature magnitude. The horizontal axis measures how each feature changes the dominant expert's routing weight. For example, high Momentum values are concentrated on the positive side, indicating that stronger recovery momentum generally reinforces the dominant-expert assignment. In contrast, for most features, including DI, PreVar, ImpPersist, and RecDef, both high and low values appear on either side of zero. It is understandable that the same feature may strengthen the dominant-expert weight for some samples but weaken it for others, revealing nonlinear and context-dependent routing behavior.

The right panel shows that impact depth has the largest mean absolute attribution (15.0\%), followed by momentum (11.7\%) and built-up area (11.2\%). Pre-event variability, impact persistence, and recovery
deficit each contribute approximately 9\%, while recovery rate, observed STP, population density, and the baseline restored ratio contribute 7.9\%,
6.2\%, 5.8\%, and 4.2\%, respectively. At the feature-group level, recovery, contextual, and impactfeatures account for 38.4\%, 31.4\%, and 30.2\% of the total attribution. Thus, all three information groups play important roles in expert assignment,with recovery features contributing slightly more overall.

For implementation, we use Captum's GradientSHAP
\cite{appendix-e2-01} with an event-balanced background of 200 training samples, 1,024 Monte Carlo samples, and Gaussian noise ($\sigma=0.01$). 95\% confidence intervals are obtained from 5,000 event-level bootstrap replicates.






\end{document}